\documentclass[sigplan,screen]{acmart}

\usepackage{amsmath}
\usepackage{enumitem}
\usepackage{algorithm}
\usepackage{multirow}
\usepackage{balance}
\AtBeginDocument{%
  }

\copyrightyear{2026}
\acmYear{2026}
\setcopyright{cc}
\setcctype{by}
\acmConference[ASPLOS '26]{Proceedings of the 31st ACM International Conference on Architectural Support for Programming Languages and Operating Systems, Volume 2}{March 22--26, 2026}{Pittsburgh, PA, USA}
\acmBooktitle{Proceedings of the 31st ACM International Conference on Architectural Support for Programming Languages and Operating Systems, Volume 2 (ASPLOS '26), March 22--26, 2026, Pittsburgh, PA, USA}
\acmPrice{}
\acmDOI{10.1145/3779212.3790223}
\acmISBN{979-8-4007-2359-9/2026/03}

\begin{document}

%%
%% The "title" command has an optional parameter,
%% allowing the author to define a "short title" to be used in page headers.
\title{SNIP: An Adaptive Mixed Precision Framework for Subbyte Large Language Model Training}

%%
%% The "author" command and its associated commands are used to define
%% the authors and their affiliations.
%% Of note is the shared affiliation of the first two authors, and the
%% "authornote" and "authornotemark" commands
%% used to denote shared contribution to the research.
\author{Yunjie Pan}
\affiliation{%
  \institution{University of Michigan}
  % \streetaddress{P.O. Box 1212}
  \city{Ann Arbor}
  \state{Michigan}
  \country{USA}
}
\email{panyj@umich.edu}

% \author{Jiecao Yu}
% \affiliation{%
%   \institution{Meta Platforms, Inc.}
%   % \streetaddress{P.O. Box 1212}
%   \city{Menlo Park}
%   \state{California}
%   \country{USA}
% }
% \email{jiecaoyu@meta.com}

\author{Yongyi Yang}
\affiliation{%
  \institution{University of Michigan}
  % \streetaddress{P.O. Box 1212}
  \city{Ann Arbor}
  \state{Michigan}
  \country{USA}
}
\affiliation{
  \institution{NTT Research, Inc.}
  \city{Sunnyvale}
  \state{California}
  \country{USA}
}
\email{yongyi@umich.edu}

\author{Hanmei Yang}
\affiliation{%
  \institution{University of Massachusetts Amherst}
  % \streetaddress{P.O. Box 1212}
  \city{Amherst}
  \state{Massachusetts}
  \country{USA}
}
\email{hanmeiyang@umass.edu}

% \author{Summer Deng}
% \affiliation{%
%   \institution{Meta Platforms, Inc.}
%   % \streetaddress{P.O. Box 1212}
%   \city{Menlo Park}
%   \state{California}
%   \country{USA}
% }
% \email{summerdeng@meta.com}

\author{Scott Mahlke}
\affiliation{%
  \institution{University of Michigan}
  % \streetaddress{P.O. Box 1212}
  \city{Ann Arbor}
  \state{Michigan}
  \country{USA}
}
\email{mahlke@umich.edu}

%%
%% By default, the full list of authors will be used in the page
%% headers. Often, this list is too long, and will overlap
%% other information printed in the page headers. This command allows
%% the author to define a more concise list
%% of authors' names for this purpose.
\renewcommand{\shortauthors}{Yunjie Pan, Yongyi Yang, Hanmei Yang, and Scott Mahlke}

\newcommand{\THISWORK}{SNIP}
\newcommand{\FrobNorm}[1]{\left\| #1 \right\|_F}

%%
%% The abstract is a short summary of the work to be presented in the
%% article.
\begin{abstract}
Training large language models (LLMs) efficiently while preserving model quality poses significant challenges, particularly with subbyte precision supported by state-of-the-art GPUs. Current mixed-precision training approaches either apply uniform precision to all GEMM operations or rely on heuristic-based methods that fail to generalize during training, leading to suboptimal convergence and instability.

To address these challenges, this paper introduces \THISWORK, a fine-grained adaptive mixed-precision training framework for LLM pretraining that supports subbyte precision. \THISWORK\ periodically collects statistics on activations, gradients, and optimizer states to assess the precision loss impact on model quality. We define two key metrics: \textit{loss divergence} in the forward pass, caused by quantization-induced increases in training loss, and \textit{weight divergence} in the backward pass, which measures error propagation through gradients affecting model updates. These metrics guide an Integer Linear Programming (ILP) problem that systematically optimizes layerwise precision to minimize overall quality loss while meeting efficiency targets. 
Experiments on 1B, 3B, 7B and 70B Llama-like models demonstrate that \THISWORK\ consistently outperforms existing baselines, reducing FLOPs by up to 80\% while preserving model quality across different model sizes and training phases with minimal computational overhead.
\end{abstract}

%%
%% The code below is generated by the tool at http://dl.acm.org/ccs.cfm.
%% Please copy and paste the code instead of the example below.
%%


%%
%% Keywords. The author(s) should pick words that accurately describe
%% the work being presented. Separate the keywords with commas.
%% A "teaser" image appears between the author and affiliation
%% information and the body of the document, and typically spans the
%% page.

% \received{20 February 2007}
% \received[revised]{12 March 2009}
% \received[accepted]{5 June 2009}

%%
%% This command processes the author and affiliation and title
%% information and builds the first part of the formatted document.
\maketitle

\section{Introduction}
% \TODO{mention H100 supports FP8, and B100 supports FP4. TFLOPS on GPUs. FP4 1.4x speedup over FP8 and FP8 has 1.4x speedup over BF16}
% \TODO{Add future work section. Memory saving for storing lower precision of outputs}
Large Language Models (LLMs) have revolutionized natural language processing (NLP), powering applications such as conversational AI, code generation, and reasoning~\cite{brown2020language,achiam2023gpt,touvron2023llama,dubey2024llama, liu2024deepseek, team2023gemini, bai2023qwen}. Despite their widespread utility, training LLMs demands extraordinary computational resources. For example, training Llama 3~\cite{dubey2024llama} with 8 billion parameters requires 1.46 million GPU hours on NVIDIA H100 GPUs, resulting in 420 tons of CO2 emissions.
The immense computational resources, runtime, and environmental impact underscore the urgent need for efficient training strategies.

Mixed precision training~\cite{micikevicius2018mixed} has emerged as a pivotal technique to mitigate the costs, allowing most compute-intensive operations to be executed in low-precision formats (e.g. FP8~\cite{liu2024deepseek,peng2023fp8,micikevicius2022fp8}, FP4~\cite{wang2025optimizing}), while retaining higher precision for critical operations. 
The NVIDIA Hopper GPU supports the FP8 format~\cite{Hopper}, while the more recent Blackwell GPU extends support to the FP4 format~\cite{Blackwell}. For the FP8 format, the GEMM operation achieves twice the TFLOPS of BF16, resulting in a 1.3x to 1.4x speedup in end-to-end LLM training. Similarly, an FP4 GEMM on the Blackwell GPU doubles the TFLOPS compared to FP8, which translates into an additional 1.4x speedup over  FP8 training latency~\cite{tseng2025training}.

%This approach effectively reduces computational costs while preserving model quality, striking a balance between efficiency and numerical stability.
However, most existing mixed precision training frameworks adopt a uniform precision policy, assuming identical precision requirements for all layers, thereby missing the opportunity for greater efficiency gains through fine-grained precision configurations~\cite{liu2024deepseek,peng2023fp8,micikevicius2022fp8,wang2025optimizing}. While prior research has investigated layer-wise quantization for machine learning models, these efforts are largely focused on inference rather than training and come with limitations.
%For example, prior works primarily focus on evaluating the \textit{local} impact of quantization using proxy metrics such as KL-divergence~\cite{wang2019haq, chen2024bbs}, absolute quantization error, or relative quantization error.
%When selecting quantization policies, prior methods often rely on computationally intensive or non-transparent approaches. For instance, HAQ~\cite{wang2019haq} uses reinforcement learning (RL) with complex reward and action spaces, OBC~\cite{frantar2022optimal} uses Hessian matrix for layer-wise pruning, and BitSET~\cite{pan2023bitset} employs heuristics to iteratively reduce precision by 1 bit per layer, followed by accuracy testing on a small dataset subset to estimate sensitivity to precision loss. Such techniques are computationally expensive as they require multiple iterations of model execution to approximate the end-to-end accuracy degradation. Similarly, BBS~\cite{chen2024bbs} determines layer sensitivity based solely on weight magnitude, oversimplifying the intricate relationship between quantization and training stability. These methods often overlook the broader implications of quantization across layers and training stages, limiting their applicability in training large-scale models like LLMs.
These methods often rely on local quantization metrics such as KL-divergence~\cite{wang2019haq, chen2024bbs}, absolute quantization error, or relative quantization error, which fail to fully capture the broader training dynamics. Additionally, some prior work employs computationally expensive approaches, including reinforcement learning (HAQ~\cite{wang2019haq}), Hessian-based sensitivity analysis (OBC~\cite{frantar2022optimal}), or iterative precision refinement (BitSET~\cite{pan2023bitset}). These techniques are primarily tailored for inference and do not address the comprehensive needs of model quality during LLM pretraining, thus limiting their effectiveness in a training context. %overlook the global impact of quantization on training quality, often resulting in suboptimal performance for large-scale LLMs.

To address these limitations, we propose \THISWORK, a fine-grained mixed precision training framework that dynamically determines the quantization per layer to ensure training quality and efficiency. Rather than relying on exhaustive search or purely heuristic-based decisions, \THISWORK\ introduces a novel optimization proxy quality metric to quantify the impact of quantization. %Furthermore, \THISWORK\ is designed to be compatible with emerging quantization techniques, as it selects layer-wise precision assignments from a set of candidate quantization schemes, ensuring continued compatibility with future advancements.

\begin{figure}
    \centering
    \includegraphics[width=0.9\linewidth]{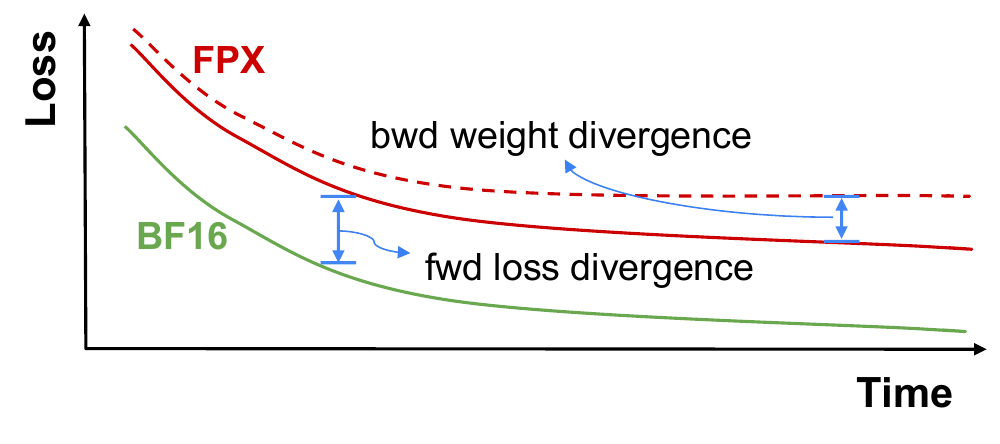}
    \caption{Illustration of the gap of the training loss between high-precision (BF16) and low-precision (FPX). The gap consists of two parts: (1) Forward Loss Divergence: the increase in training loss directly introduced by quantization during the forward pass, and (2) Backward Weight Divergence: the accumulation of errors in weight updates during the backward pass due to quantization, which can compound over iterations and layers, impacting model convergence.}
    \label{fig:loss_diagram}
\end{figure}

\THISWORK\ quantifies training quality loss through two key metrics, as illustrated in Figure~\ref{fig:loss_diagram}: loss divergence and weight divergence. Loss divergence occurs in the forward pass when quantization increases training loss. Weight divergence occurs in the backward pass when quantization errors in gradients distort parameter updates, resulting in the trained model deviating from the models trained with full precision. %. Unlike inference, where quantization primarily affects output activations, training involves iterative updates where errors can accumulate across layers and iterations, leading to instability or degraded convergence. 
Building upon these metrics, we formulate the problem of determining layer-specific quantization schemes as an integer linear programming (ILP) problem.  The ILP systematically minimizes quality loss while meeting efficiency constraints, ensuring a globally optimized quantization policy.

\begin{figure}
    \centering
    \includegraphics[width=0.95\linewidth]{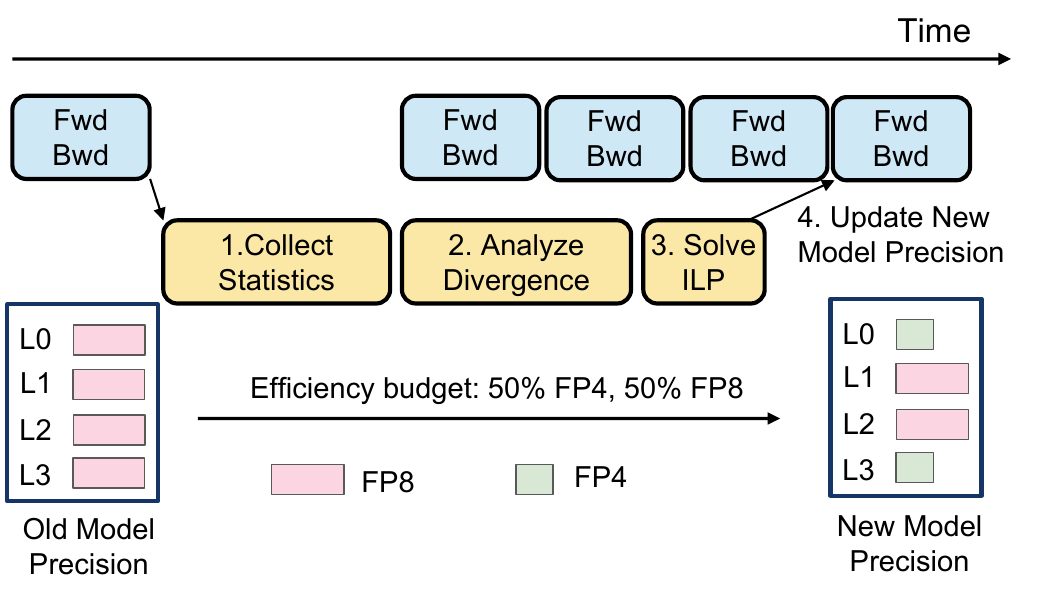}
    \caption{System overview of \THISWORK, showing its integration into LLM training. \THISWORK\ periodically collects statistics on activations, weights, and optimizers, then asynchronously analyzes divergence metrics, solves an ILP problem, and updates layer-wise quantization.}
    \label{fig:systme_simplify}
\end{figure}

\begin{figure}
    \centering
    \includegraphics[width=\linewidth]{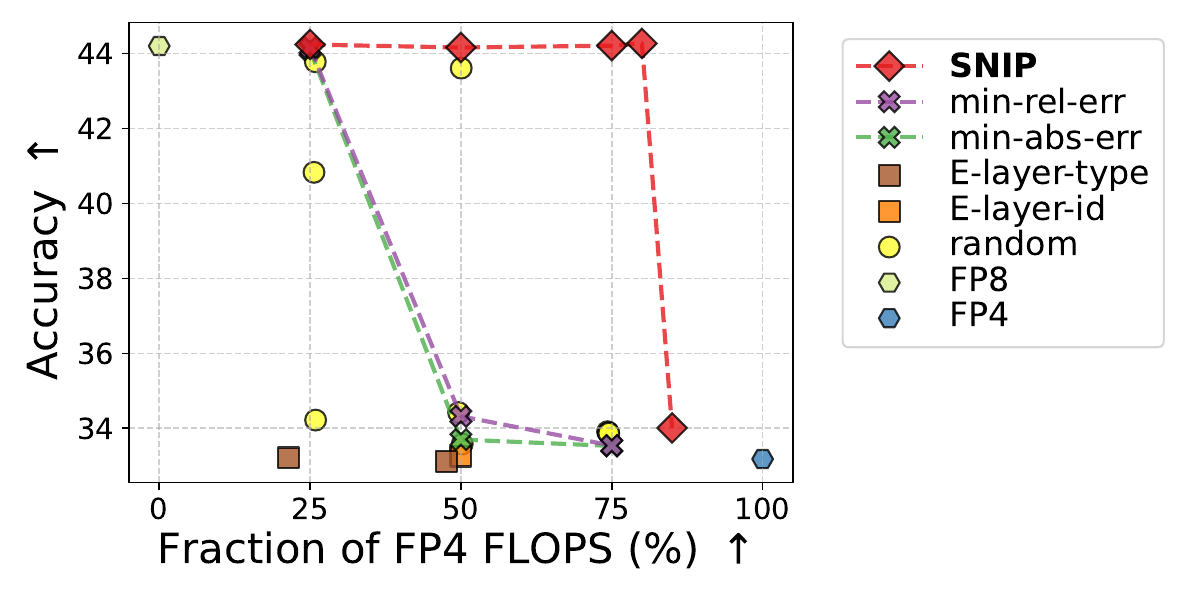}
    \caption{Comparison of accuracy versus efficiency (the fraction of FP4 FLOPs) for the TinyLlama 1B model. The FP8 baseline achieves the highest accuracy but lowest efficiency, while the FP4 configuration maximizes efficiency at the cost of accuracy. \THISWORK\ demonstrates a balance between high accuracy and a significant reduction in FLOPs by selectively applying FP4 to certain layers based on the loss and weight divergence metrics, outperforming other methods such as \textit{random} and other heuristic-based approaches.}
    \label{fig:effi_acc_highlight}
\end{figure}

Beyond its theoretical advancements, \THISWORK\ designs a practical system that integrates seamlessly into LLM training pipelines, as shown in Figure~\ref{fig:systme_simplify}. The system operates asynchronously alongside standard training, periodically updating layer-wise quantization decisions without interrupting the primary training loop. The system follows a structured workflow: (1) collect statistics on activations, gradients, and optimizer states, (2) analyze these metrics to estimate the impact of precision scaling, (3) formulate an optimization problem to determine the optimal per-layer precision settings, and (4) update quantization assignments asynchronously. This adaptive approach ensures that \THISWORK\ remains effective across different training phases and model architectures, making it broadly applicable to various LLM workloads.

Figure~\ref{fig:effi_acc_highlight} highlights the trade-off between the accuracy and efficiency of different precision selection methods for the TinyLlama 1B model. Efficiency is measured as the fraction of FLOPs executed in FP4, with the remainder using FP8 precision. \THISWORK\ consistently outperforms all other quantization methods at different efficiency levels, including \textit{random} (randomly selecting layers for FP4), \textit{min-abs-err} (selecting layers that minimize absolute quantization error for FP4), \textit{min-rel-err} (selecting layers that minimize relative quantization error for FP4), \textit{E-layer-type} (empirically apply FP4 to non-sensitive layer types), and \textit{E-layer-id} (empirically apply FP4 for middle layers). Remarkably, even at 80\% FP4 FLOPs, \THISWORK\ maintains nearly full-precision accuracy, whereas alternative quantization methods fail to converge.

The contributions of this paper are:
\begin{itemize}[leftmargin=*]
    \item We propose \THISWORK\, a fine-grained mixed-precision quantization framework specifically for LLM pretraining that seamlessly integrates into existing pipelines. \THISWORK\ periodically determines the optimal quantization policy while minimizing the model quality loss. Unlike heuristic-based methods, \THISWORK\ provides a global optimization strategy that dynamically adapts to different training stages and supports various quantization techniques. %This framework is designed to handle the evolving nature of the training process effectively, balancing precision and efficiency without compromising training quality.
    \item We present a novel theoretical perspective on how quantization errors influence overall LLM training quality. 
    Specifically, we introduce two quantization quality metrics, loss divergence for the forward pass and weight divergence for the backward pass, which quantify the impact of precision scaling on training stability. These metrics enable efficient precision selection with minimal overhead. 
    %Specifically, we propose two novel quantization quality metrics—loss divergence (forward pass) and weight divergence (backward pass)—that quantify the impact of precision scaling on training stability. These metrics enable efficient precision selection with minimal overhead.
    \item We evaluate \THISWORK\ on 1B, 3B, 7B, and 70B models using up to 80\% FP4 FLOPs, demonstrating that it consistently improves training efficiency with subbyte precision while maintaining near full-precision model accuracy.  %Yunjie emphasize FP4 results here.
\end{itemize}
\section{Background}
\subsection{LLM Structure}
\begin{figure}
    \centering
    \includegraphics[width=0.95\columnwidth]{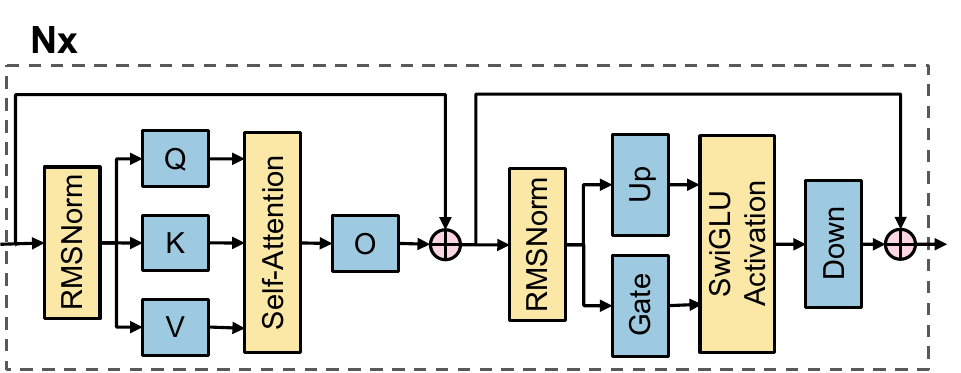}
    \caption{The transformer block structure of Llama-like LLM. The blocks in blue (Q, K, V, O, Gate, Up and Down) are linear layers.}
    \label{fig:transformer_block}
\end{figure}

\begin{figure}
\centering
\includegraphics[width=\columnwidth]{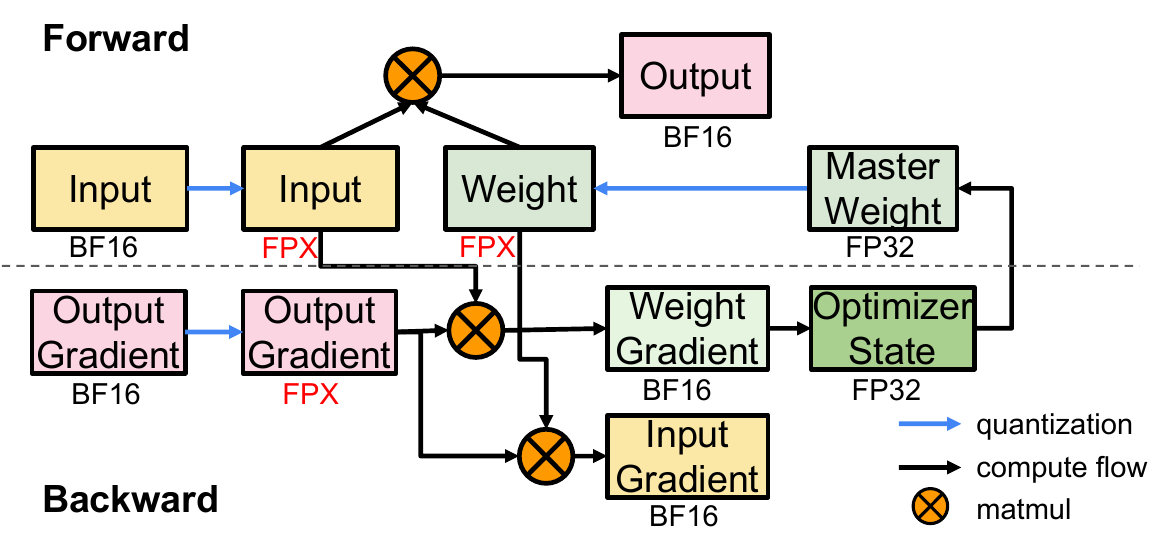}
\caption{The overall mixed precision training framework for linear layers, including the forward and backward pass.}
\label{fig:mixed_precision_training}
\end{figure}
Most LLMs are built upon the Transformer~\cite{vaswani2017attention} framework, which consists of embedding layers that convert tokenized inputs into dense vector representations, transformer blocks that are the core computation units, stacked $N$ times to increase the model capacity, and output projection layer that maps the final hidden representations of tokens back to the vocabulary space.
We illustrate the transformer block structure of Llama-like LLMs in Figure~\ref{fig:transformer_block}. Each transformer block contains multi-head self-attention, which uses three types of linear layers Query (Q), Key (K), and Value (V) to project inputs to an intermediate representation. After computing the attention scores, the output is projected back to the original dimension using the Output linear layer (O). Subsequently, the processed tensors are passed through a Feedforward Neural Network (FFN), which comprises the Gate and Up linear layers for intermediate transformations. After the non-linear activation (e.g. SwiGLU), the intermediate representation is projected back by the Down linear layer.
Following prior works~\cite{liu2024deepseek, wang2025optimizing}, we focus on the quantization of those linear layers in transformer blocks because they take the majority (>90\%) of the FLOPs during training~\cite{transformerFlops}. %they are the most compute intensive parts of the model during training.

\begin{figure*}
    \centering
    \includegraphics[width=0.8\linewidth]{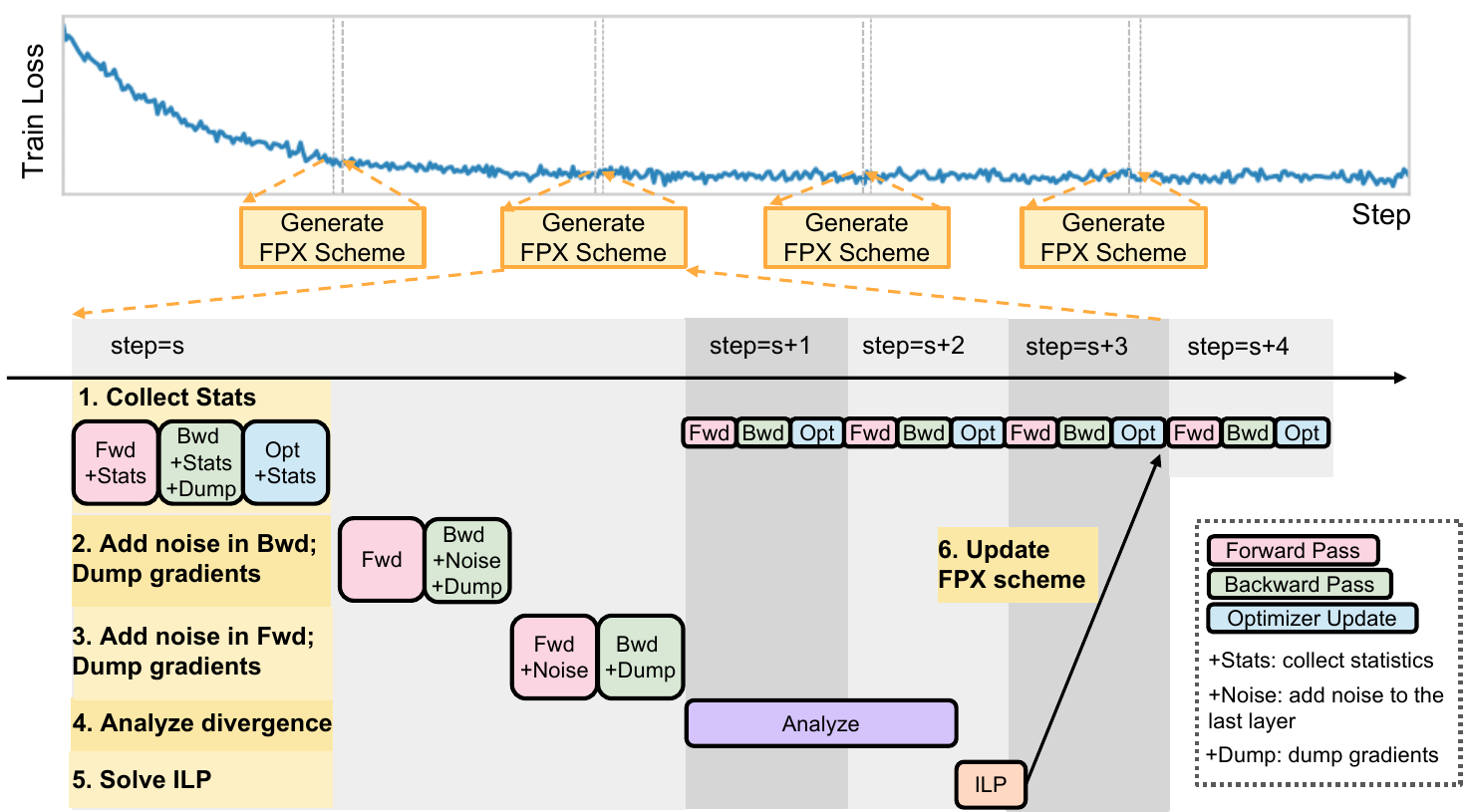}
    \caption{An overview of the \THISWORK\ training workflow. The top plot illustrates the training loss over steps, with periodic updates to the FPX quantization scheme highlighted in yellow boxes. The process consists of six steps: (1) collecting statistics during a normal iteration, (2) injecting noise during the backward pass and dumping gradients, (3) injecting noise during the forward pass and dumping gradients, (4) analyzing loss and weight divergence, (5) solving the Integer Linear Programming (ILP) problem to determine the optimal FPX quantization scheme, and (6) updating the FPX scheme.}
    \label{fig:overall}
\end{figure*}
\subsection{Mixed Precision Training}
Mixed precision training~\cite{micikevicius2018mixed} addresses training costs and model accuracy by using reduced precision, typically BF16, in linear or convolutional layers while retaining higher precision for critical operations. %These include maintaining a master copy of weights in high precision and accumulating low-precision arithmetic results in high precision for stability.
State-of-the-art advancements such as DeepSeek-V3~\cite{liu2024deepseek} have extended these techniques to include FP8 training for large language models (LLMs). Meanwhile, Wang \textit{et al.}~\cite{wang2025optimizing} have integrated FP4 into mixed precision training. %for inputs and weights, while leaving the gradients in higher precision, where 2 out of 3 matrix multiplication in training involves gradients. 
The FP4 approaches, while employing FP4 across all layers, necessitate complex gradient calculations that add overhead, and rely on irregular sparse GEMM to handle outliers, which further adds significant overhead. %, achieving minimal loss error compared to the BF16 baseline. This work highlights the effectiveness of leveraging low-precision formats to balance computational efficiency and model quality during training.

Adopting a similar idea on mixed precision training for LLMs, we present the overall framework used in our experiments in Figure~\ref{fig:mixed_precision_training}. In this framework, the most compute-intensive operators (GEMM) within linear layers are executed in low precision (\textit{FPX}), where \textit{FPX} refers to low-precision formats such as FP8 or FP4. Before the GEMM operation, the input activations, weights, and output gradients are quantized to low precision, while the output of the GEMM operator remains in BF16. To ensure numerical stability, we maintain a master copy of the weights in FP32, following DeepSeek-V3~\cite{liu2024deepseek}.
% Our experiment verified that optimizers using BF16 will lead to severe training divergence. 
Other types of layers, such as RMSNorm, SwiGLU, Softmax, and Attention, continue to use higher precision (BF16) to preserve accuracy while optimizing overall efficiency. This hybrid approach ensures a balanced trade-off between computational cost and model performance. 

Mixed precision training with low-precision formats such as FP4 and FP8 reduces both computation and communication overhead. On NVIDIA Blackwell~\cite{Blackwell}, FP4 offers 2$\times$ the throughput of FP8 and 4$\times$ that of BF16. 
Besides, storing weights in FP4/FP8 also reduces HBM storage cost, which is the main bottleneck in large-scale LLM training. For communication, pipeline parallelism ensures that FSDP communication is not on the critical path for the training. Extending low-precision support to reduce-scatter is a promising but challenging direction for future work. %tensor synchronization (e.g., all gather) incur less bandwidth, further reducing communication cost.}

\subsection{Quantization}
\noindent\textbf{Quantization Format.}
% Due to the dynamic range requirements of training, prior works use floating point rather than integers to represent inputs, weights and gradients. 
Common floating point types for training include IEEE single precision (FP32) and bfloat16 (BF16)~\cite{kalamkar2019study}. To reduce the number of bits in floating point further, FP8 format has been proposed and integrated into commercial GPUs. There are different FP8 formats studied~\cite{micikevicius2022fp8, shen2024efficient}: E5M2, E4M3 and E3M4, where the names indicate the number of exponent (E) and mantissa (M) bits, respectively. Following MX specification~\cite{rouhani2023microscaling}, we adopt the FP4 E2M1 format.

\noindent\textbf{Quantization granularity.}
Because low-precision formats (e.g., FP8 and FP4) have a limited dynamic range, scaling factors are essential to mitigate numerical overflows and underflows. This is particularly crucial during training, as gradients often exhibit a larger dynamic range~\cite{liu2024deepseek}, and their numerical accuracy significantly affects model convergence.
As a standard practice, the input distribution is aligned to the representable range of the FPX format by scaling the maximum absolute value of the input tensor to the maximum representable value of FPX ($FPX\_{MAX})$. 
$$
scale = FPX\_{MAX} / max(abs(x))
$$
Before applying the low-precision operator ($Op$), the inputs are scaled by the scaling factor and then quantized to the low-precision format. After the operator, the output is scaled back by dividing by the same scaling factor.
$$
y = Op(Quant(x * scale)) / scale
$$

To reduce quantization error, tensors can be scaled at different levels of granularity: (1) tensorwise: Every tensor has a scaling factor; (2) rowwise/columnwise: Every row or column of a tensor has a scaling factor; (3) blockwise: Every $N_B\text{x}N_B$ block of a tensor has a scaling factor; (4) tilewise: Every $1\text{x}N_B$ tile of a tensor has a scaling factor. 
Following DeepSeek-V3~\cite{liu2024deepseek}'s strategy, we apply $1\text{x}128$ tilewise quantization for input activations and gradients, and $128\text{x}128$ blockwise quantization for weights.

\section{Overall Procedure}

In this section, we present the overall process for generating layer-wise quantization schemes (FPX schemes) during the pretraining of LLMs, as illustrated in Figure~\ref{fig:overall}. Throughout the training process, optimal quantization schemes are periodically generated and applied asynchronously, ensuring seamless integration with the normal training workflow. 
%We update quantization schemes periodically to ensure efficiency and stability. First, value distributions remain stable over short training intervals, making frequent updates unnecessary. 
%Empirically, updating every 50k to 100k steps is sufficient. \hanmei{is there any data or references to prove it?} Second, generating quantization schemes incurs computational overhead from collecting statistics and solving optimization problems. Periodic updates balance this cost while maintaining training performance.
% We generate quantization schemes periodically for two key reasons: 1) Stability of Value Distribution: The value distribution over a relatively short training time frame tends to remain consistent. Therefore, there is no need to perform continuous online updates to the quantization scheme. Empirically, we found that updating the quantization scheme every 100k steps is sufficient; 2) Minimizing Overhead: The process of generating quantization schemes involves additional steps, such as collecting statistics and solving optimization problems, which introduces some computational overhead. Periodic updates help balance this overhead while maintaining training efficiency.

Specifically, the generation of FPX schemes consists of two main tasks: (a) Collecting Statistics and Analyzing Divergence; (b) Determining the optimal layer-wise quantization scheme. Task (a) comprises Steps 1 through 4, while Task (b) includes Steps 5 and 6. We use the example of Figure~\ref{fig:overall} that collects the statistics regarding the checkpoint at step $s$ to generate the optimal quantization scheme.

\subsection{Collecting Statistics and Analyzing Divergence}
In Step 1 of Figure~\ref{fig:overall}, we collect statistics during a standard training iteration using high precision (BF16), which includes the forward pass, backward pass, and optimizer updates, while also dumping gradient tensors. 
Specifically, during the forward and backward passes, we record key statistics such as the Frobenius norm of inputs, weights, outputs, output gradients, input gradients, and weight gradients. Additionally, we compute and store the Frobenius norm of the quantization error for these components. During the optimizer update, we gather statistics like the first and second moments to accurately model weight update dynamics. Since most statistics involve simple Frobenius norm computations, the overhead of collecting statistics is negligible compared to a standard training iteration. The theoretical justification for collecting this information is discussed in detail in Section~\ref{sec:quantify}.

Steps 2 and 3 approximate the Frobenius norm of the second-order derivative to estimate the impact of quantization on the backward weights update, as calculating its exact value is computationally prohibitive. The theoretical rationale behind these steps is discussed in detail in Section~\ref{sec:quantify_bwd}. In Step 2, using the same batch of data from training step $s$, we inject small Gaussian noise into the last layer during the backward pass and save the gradient tensors. Later, these gradients are compared with the baseline gradients (collected in Step 1 without noise) by computing the Frobenius norm of their difference. Similarly, in Step 3, Gaussian noise is added to the last layer during the forward pass, and the resulting gradients are compared with the baseline gradients.
Step 2 and 3 require two additional forward and backward passes without updating the weights that is done on GPU, so these 2 steps introduce some computational overhead.

In Step 4, we leverage the collected statistics and dumped tensors to analyze and compute two key metrics: loss divergence, which captures the increase in training loss caused by quantization errors during the forward pass, and weight divergence, which quantifies the deviation of the weights from the baseline model due to quantization errors. This analysis can be offloaded to the CPU, allowing the normal training process to continue seamlessly with subsequent steps, as illustrated in the example for steps $s+1$ and $s+2$ in Figure~\ref{fig:overall}.

Our algorithm is agnostic to the parallelism strategy and requires only minor implementation considerations. The 70B experiments in Section~\ref{sec:evaluation} confirm that it works seamlessly under different parallelism configurations. % With fully sharded data parallelism (FSDP), parameters are gathered before computation, so sharding does not alter the algorithm. Under tensor parallelism (TP), statistics are first collected on sharded tensors and then logically concatenated to recover full-tensor information. For pipeline parallelism (PP), layers are distributed across GPUs but tensors are not sharded, so the algorithm remains unaffected.

\subsection{Deciding the Optimal Layer-wise Quantization Scheme}

In Step 5, the loss and weight divergence are used to formulate an Integer Linear Programming (ILP) problem. The goal of this optimization is to identify the optimal quantization scheme for each layer, balancing training quality and efficiency requirements.
The ILP maps the problem to a knapsack model, where each layer is a decision variable, and precision formats (e.g., FP8, FP4) are quantization options.  \THISWORK\ is compatible with emerging quantization techniques, as new methods can be incorporated as additional quantization options. The objective function minimizes a weighted combination of the total loss divergence and weight divergence across all layers, subject to efficiency constraints (e.g., the fraction of FLOPs allocated to FP4 precision).

After solving the ILP problem in Step 5, the optimal quantization scheme is applied to the training process in Step 6. To minimize overhead and maintain training efficiency, the updated scheme is applied asynchronously without interrupting the ongoing training iterations. The updated scheme remains in effect until the next quantization update cycle. %, typically occurring every 50k-100k iterations.

% These steps collaboratively generate and implement optimal quantization schemes at regular intervals, striking a balance between training efficiency and model quality with minimal disruption to the normal training process.

% analyzing the quantization impact and using heuristics to decide the optimal layer-wise quantization scheme for mixed precision training in LLMs.

% \begin{algorithm}
% \caption{Fined-grained mixed precision training}
% \label{alg:fine_grained_precision}
% \begin{flushleft}
% \textbf{Input:} Model parameters $\Theta$, dataset $\mathcal{D}$, training steps $T$, interval for precision update $n$, efficiency target $E_t$ \\
% \textbf{Output:} Trained model $\Theta$
% \end{flushleft}
% \begin{algorithmic}[1]
% \State Initialize precision settings for all layers with baseline (e.g., FP8)
% \For{$t = 1$ to $T$}
%     \State Perform forward and backward passes with current precision settings
%     \State Update model parameters $\Theta$ using optimizer (e.g., AdamW)
%     \If{$t \% n == 0$} \Comment{Every $n$ steps}
%         \State Collect statistics for activations, weights, and gradients
%         \State Analyze \textit{loss divergence} ($\Delta L$) and \textit{weight divergence} ($\Delta W$)
%         \State Formulate and solve MILP for optimal layer-wise precision settings 
%         \State Update precision settings for each layer based on MILP results
%     \EndIf
% \EndFor
% \State \Return $\Theta$
% \end{algorithmic}
% \end{algorithm}
\section{Quantify the Quantization Impact}
\label{sec:quantify}
\subsection{Preliminary}
Figure~\ref{fig:mixed_precision_training} shows the quantization process in LLM mixed precision training. We only consider linear layers in transformer blocks, whose computation is matrix multiplication.

Quantization reduces the number of bits used to present the tensors of inputs, weights, and gradients in DNNs, that incurs error. Following prior work~\cite{bengio2013estimating, lee2016training}, we model the impact of quantization of a tensor as a small random perturbation term. Specifically, 
$$
q(x) = x + \delta_x
$$
where $x$ is the ground truth value of a tensor, $q(x)$ is the quantized value of $x$, and $\delta_x$ is a small random Gaussian vector.

We denote $\FrobNorm{\cdot}$ as the Frobenius norm %\yyy{better use $\|\cdot\|_F$ for matrix as $\|\cdot\|$ usually stands for L2 norms}
for a matrix (equivalent to the \(\ell_2\) norm of the vectorization of a matrix) and the \(\ell_2\) norm for a vector. We use $I_d$ to represent the identity matrix of size $d \times d$, and $\mathcal N$ to represent the Gaussian distribution. Below we prove two theorems that are critical to our analysis. The idea behind these theorems is that if the perturbation is random enough, then the error can be estimated much more accurately than a trivial one (say, using the Lipschitz constant). Similar ideas have been adopted in recent work on quantization under different settings~\cite{ext-rabitq,raana}.  
%%% I currently comment it out in case the work is not on arxiv when I submit this paper.
%This idea of Theorem~\ref{thm:perturb} is similar with a concurrent work RabitQ\yyy{cite varied rabitq} but under a slightly different setting.\yyy{I'm still not sure about the exact name of varied rabitq, perhaps can skip it for you early version.}
\begin{theorem}\label{thm:perturb}
%\yyy{I assume $\delta$ is Gaussian here. If $\delta$ is Uniform on sphare then the proof would be slightly more complicated.} 
Let $\epsilon > 0$ be a small scalar, $\delta_x \sim \mathcal N\left(0, \frac{\epsilon^2}{d}I_d\right)$ be a random Gaussian vector, and $g: \mathbb R^d \to \mathbb R^m$ be a smooth function with $S$-Lipschitz gradient such that $S \ll \frac{1}{\epsilon^2}$, then the following inequality holds with probability at least $0.99$:\begin{align*}
\left\| g(x+\delta_x) -  g(x)\right\|_F \lesssim \left\|\nabla g(x)\right\|_F \frac{\epsilon}{\sqrt{d}},
\end{align*}
where we use $\lesssim$ to hide constant coefficients. 

    % Let $\delta \in \mathbb R^{d}$ be a small random perturbation vector with zero-mean independent entries, where $\FrobNorm{\delta} \leq \epsilon$ for some small $\epsilon$. Suppose $g: \mathbb R^{d} \to \mathbb R $ is a smooth function with Lipschitz-continuous gradient, then with high probability ? 
    % $$
    % \FrobNorm{ g(x+\delta) - g(x)} =  \FrobNorm{\nabla_x g(x)}  \FrobNorm{ \delta} / \sqrt{d}
    % $$
\end{theorem}

Theorem \ref{thm:perturb} illustrates how we can use the gradient norm to estimate the error under a small perturbation. Conversely, we can also use the error under perturbation to estimate the gradient norm.

\begin{theorem}\label{thm:est-gradient-norm} 
Let $g: \mathbb R^d \to \mathbb R^m$ be a smooth function, and let $\delta_x \sim \mathcal N(0, \epsilon^2 I_d)$ be a standard Gaussian vector, then we have \begin{align*}
 \|\nabla _x g(x)\|_F^2 = \lim_{\epsilon \to 0} \epsilon^{-2} \mathbb E \|g(x) - g(x +  \delta_x)\|_F^2. \label{eq:est-grad-norm}
\end{align*}

\end{theorem}

\subsection{Loss Divergence in Forward Pass}
\label{sec:loss_divergence_fwd}
During the forward pass, quantizing activations and weights introduces perturbations that propagate through subsequent layers, ultimately affecting the loss. For a given layer $l$, the loss can be represented as a $L(X_l, W_l)$, where $X_l \in \mathbb R^{M_l\times K_l}$ and $W_l \in \mathbb R^{N_l\times K_l}$ are the activations and weights of layer $l$. 
Using Theorem~\ref{thm:perturb}, we estimate the impact of quantizing a tensor by the following approximations:
{\small \begin{align*}
 & \left\| L(X_l+\delta_{X_l}, W_l + \delta_{W_l}) - L(X_l, W_l) \right\|
 \\ \approx &  \sqrt{ \left\| L(X_l+\delta_{X_l}, W_l) - L(X_l, W_l) \right\|^2 
  + \left\| L(X_l, W_l+\delta_{W_l}) - L(X_l, W_l) \right\|^2},
\end{align*}}
where
\begin{align*}
\FrobNorm{L(X_l+\delta_{X_l}, W_l) - L(X_l, W_l)}  \approx \frac{\FrobNorm{\nabla_{X_l} L} \FrobNorm{\delta_{X_l}} }{\sqrt{M_lK_l}}
\end{align*}
and
\begin{align*}
\FrobNorm{L(X_l, W_l + \delta_{W_l}) - L(X_l, W_l)}  \approx \frac{\FrobNorm{\nabla_{W_l} L} \FrobNorm{\delta_{W_l}} }{\sqrt{N_lK_l}}.
\end{align*}
Notice that we omit the cross term $L(X_l + \delta_{X_l}, W_l + \delta_{W_l}) - L(X_l,W_l)$ here, since it is of magnitude $O(\|\delta_{X_l}\|\|\delta_{W_l}\|)$ which is small compared to the other terms.

Here, $\nabla_{X_l} L$ and $\nabla_{W_l} L$ are the gradient of $L$ with respect to activation $X_l$ and weights $W_l$, respectively. These gradients are naturally computed during the backward pass of LLM training, allowing us to quantify the impact of quantization on loss with minimal computational overhead in Step 1 of Figure~\ref{fig:overall}.

We define the normalized loss divergence as follows.
\begin{definition}
\label{def:Delta_L}
Normalized loss divergence $\Delta L = |L'-L|/|L|$, where $L$ and $L'$ are the loss of the model during forward pass without and with quantization, respectively. 
\end{definition}

\subsection{Weight Divergence in Backward Pass}
\label{sec:quantify_bwd}
\subsubsection{Impact on Weight Gradient}
In the backward pass, the quantization error of a given layer affects the weight gradient of that layer as well as all preceding layers.
Consider quantizing the input activation $X_l \in \mathbb R^{M_l\times K_l}$, weights $W_l \in \mathbb R^{N_l\times K_l}$, and output gradient $\nabla_{Y_l} L \in \mathbb R^{M_l\times N_l}$ of layer $l$. The weight gradient at layer $l$, denoted as $\nabla_{W_l} L$, can be represented as $g_{l}(X_j, W_j, \nabla_{Y_j} L)$, for any $j \geq l$. %\yyy{the notation is kind of loose here. Better use $g_{j,l}$ instead of $g_l$ if I understood the meaning of $g_l$ here correctly}
This implies that the weight gradient at layer $l$ is influenced not only by the quantization at the current layer but also by quantization errors in all subsequent layers. We estimate the error by the following:
{\small \begin{align*}
\FrobNorm{g_{l}(X_j + \delta_{X_j}, W_j, \nabla_{Y_j} L) - g_{l}(X_j, W_j, \nabla_{Y_j} L)}  \approx \frac{\FrobNorm{\nabla_{X_j}g_l}\FrobNorm{\delta_{X_j}}}{\sqrt{M_j K_j}},
\end{align*}}
and
{\small \begin{align*}
\FrobNorm{g_{l}(X_j, W_j + \delta_{W_j}, \nabla_{Y_j} L) - g_{l}(X_j, W_j, \nabla_{Y_j} L)}  \approx \frac{\FrobNorm{\nabla_{W_j}g_{l}}\FrobNorm{\delta_{W_j}}}{\sqrt{N_j K_j}}.
\end{align*}}

Note that $\FrobNorm{\nabla_{X_j}g_l} = \FrobNorm{\frac{\partial(\frac{\partial L}{\partial W_l})}{\partial X_j}}$ and $\FrobNorm{\nabla_{W_j}g_l} = \FrobNorm{\frac{\partial(\frac{\partial L}{\partial W_l})}{\partial W_j}}$ are second-order partial derivatives, which is not computed during the forward or backward pass in LLM training. To estimate $\FrobNorm{\nabla_{X_j}g_l}$ and $\FrobNorm{\nabla_{W_j}g_l}$, we apply Theorem~\ref{thm:est-gradient-norm}.  Here, we approximate the expectation by a single sample per batch and the limit by taking a very small value. This estimation requires additional backward passes with added noise, shown as Step 2 and 3 in Figure~\ref{fig:overall}.

\subsubsection{Impact on Weight Divergence}
In the optimizer update step, the quantization error of gradients $g_l = \nabla_{W_l} L$ propagates into the model parameters $W_l$.
We formally define the normalized weight divergence as follows.
\begin{definition}
\label{def:Delta_W}
$\Delta W = \sum_l \frac{1}{N}(\FrobNorm{W_l' - W_l} / \FrobNorm{W_l})$, where $W_l$ and $W_l'$ represent the weights at layer $l$ after the optimizer updates the model parameters without and with quantization, respectively. And $N$ is the number of layers.
\end{definition}

For this analysis, we focus on the AdamW optimizer~\cite{loshchilov2017decoupled}, the most widely adopted optimizer in LLM training~\cite{touvron2023llama,dubey2024llama,liu2024deepseek,bai2023qwen}. Without loss of generality, our method applies to any differentiable optimizer. For simplification, we focus on a specific layer $l$, omitting the layer subscript $l$ in subsequent expressions. The AdamW optimizer updates parameters at each step $t$ as follows:

\begin{align*}
    % g_t = \nabla_W f(W_{t-1}) \\
    W_{t-1} &\gets W_{t-1} - \alpha \lambda W_{t-1} \\
    m_t &\gets \beta_1 m_{t-1} + (1 - \beta_1) g_t \\
    v_t &\gets \beta_2 v_{t-1} + (1 - \beta_2) g_t^2 \\
    \hat{m}_t &\gets \frac{m_t}{1 - \beta_1^t} \\
    \hat{v}_t &\gets \frac{v_t}{1 - \beta_2^t} \\
    W_t &\gets W_{t-1} - \alpha \frac{\hat{m}_t}{\sqrt{\hat{v}_t} + \epsilon}
\end{align*}

where $\alpha$ is the learning rate , $\beta_1$, $\beta_2$ are exponential decay rates, $\lambda$ is weight decay, and $\epsilon > 0$ is a small constant for numeric stability, and $g_t$ is the gradients of weights at step $t$.

% \begin{algorithm}
% \caption{AdamW Optimizer}
% \label{algo:adamw}
% \begin{algorithmic}[1]
% \Require Learning rate $\alpha$, exponential decay rates $\beta_1$, $\beta_2$, weight decay $\lambda$, small constant $\epsilon > 0$, initial parameters $W_0$, objective function $f(W)$
% \State Initialize time step $t \gets 0$
% \State Initialize 1st moment vector $m_0 \gets 0$
% \State Initialize 2nd moment vector $v_0 \gets 0$
% \While{not converged}
%     \State $t \gets t + 1$
%     \State Compute gradients $g_t = \nabla_W f(W_{t-1})$
%     \State $W_{t-1} \gets W_{t-1} - \alpha \lambda W_{t-1}$
%     \State $m_t \gets \beta_1 m_{t-1} + (1 - \beta_1) g_t$
%     \State $v_t \gets \beta_2 v_{t-1} + (1 - \beta_2) g_t^2$
%     \State $\hat{m}_t \gets \frac{m_t}{1 - \beta_1^t}$
%     \State $\hat{v}_t \gets \frac{v_t}{1 - \beta_2^t}$
%     \State $W_t \gets W_{t-1} - \alpha \frac{\hat{m}_t}{\sqrt{\hat{v}_t} + \epsilon}$
% \EndWhile
% \end{algorithmic}
% \end{algorithm}

Let us define $W_t$ and $W_t'$ as the weights (model parameters) without and with quantization in the backward pass, respectively.
$$
W_t - W_t'
= (1 - \alpha \lambda) (W_{t-1} - W_{t-1}') + \alpha \left(\frac{\hat{m}_t'}{\sqrt{\hat{v}_t'}+ \epsilon} - \frac{\hat{m}_t}{\sqrt{\hat{v}_t}+ \epsilon}\right)
$$
The first term $(1 - \alpha \lambda)(W_{t-1} - W_{t-1}')$ represents the accumulated deviation from previous iterations due to quantization, which compounds over time. So we focus only on the second term, which captures the quantization effect at the current iteration $t$.

\begin{align*}
&\alpha \left(\frac{\hat{m}_t'}{\sqrt{\hat{v}_t'}+ \epsilon} - \frac{\hat{m}_t}{\sqrt{\hat{v}_t}+ \epsilon}\right) \\
= & \alpha \frac{\sqrt{1-\beta_2^t}}{1-\beta_1^t}\left(\frac{m_t'}{\sqrt{v_t'} + \epsilon} - \frac{m_t}{\sqrt{v_t} + \epsilon}\right)
% \alpha \frac{\sqrt{1-\beta_2^t}}{1-\beta_1^t} (\frac{\beta_1 m_{t-1}' + (1-\beta_1) g_t'}{\sqrt{\beta_2 v_{t-1}' + (1-\beta_2) g_t'^2}+ \epsilon} - \frac{\beta_1 m_{t-1} + (1-\beta_1) g_t'}{\sqrt{\beta_2 v_{t-1} + (1-\beta_2) g_t'^2}+ \epsilon})
\end{align*}

Within the same iteration, the values of $\alpha$, $\beta_1$, $\beta_2$, and $\delta$ remain consistent across all layers. Notably, the norms $\FrobNorm{m_t' - m_t}$ and $\FrobNorm{v_t' - v_t}$ are much smaller than $\FrobNorm{g_t' - g_t}$, as $(1-\beta_1)$ and $(1-\beta_2)$ are typically very small (often < 0.1). Therefore, we consider the following function of gradient $g$ to represent the weight divergence due to quantization: %\yyy{the symbol $g$ was used before as the gradient...}: 
\begin{align*}
    h(g) = \alpha \frac{\sqrt{1-\beta_2^t}}{1-\beta_1^t}\frac{m_t}{\sqrt{v_t} + \epsilon}
\end{align*}
For simplicity, we focus only on iteration $t$ and omit the subscript $t$ from $g_t$.

Using the Theorem~\ref{thm:perturb}, we have that 
{\small \begin{align*}
& \FrobNorm{h(g+\epsilon_g) - h(g)} \\
% & \approx \frac{\FrobNorm{\nabla_g h(g)} \FrobNorm{\epsilon_g}}{\sqrt{NK}} \\
& \approx \alpha \frac{\sqrt{1-\beta_2^t}}{1-\beta_1^t}\FrobNorm{\frac{1-\beta_1}{\sqrt{v_t}+\epsilon} - \frac{(1-\beta_2)m_t g_t}{\sqrt{v_t}(\sqrt{v_t}+\epsilon)^2}} \frac{\FrobNorm{\epsilon_g}}{\sqrt{NK}}
\end{align*}}

For implementation, in Step 1 of Figure~\ref{fig:overall}, we collect statistics during a standard training iteration using the AdamW optimizer. These include $\beta_1$, $\beta_2$, and the Frobenius norm of the term $\frac{1-\beta_1}{\sqrt{v_t}+\epsilon} - \frac{(1-\beta_2)m_t g_t}{\sqrt{v_t}(\sqrt{v_t}+\epsilon)^2}$. %In Step 4, once $\FrobNorm{\epsilon_g}$ is calculated, we can accurately quantify the impact of layerwise quantization on weight divergence.
\section{Find Optimal Quantization Policy}
\label{sec:optimization}
Given the estimated impact on quality loss and efficiency improvement, we now introduce the method to determine the optimal quantization policy that satisfies efficiency constraints while minimizing quality loss.

\subsection{Quality Loss and Efficiency Metrics}
To formulate the problem, we quantitatively define the metrics for quality loss $Q$ and efficiency $E$. 
For quality loss $Q$, we used a weighted sum of the normalized loss divergence $\Delta L$ (defined in Definition~\ref{def:Delta_L}) in the forward pass and the normalized weight divergence $\Delta W$ (defined in Definition~\ref{def:Delta_W}) in the backward pass. Then $Q = \Delta L + \Delta W$. 

For efficiency $E$, since we do not have access to GPUs that natively support both FP8 and FP4 formats, we cannot directly measure the end-to-end performance improvement of low-precision formats compared to full precision (BF16). 
%DeepSeek-V3~\cite{liu2024deepseek} demonstrates the training stability of LLMs using the FP8 format with tilewise and blockwise quantization granularity. \hanmei{The transition appears abrupt, and it would be helpful to clarify how FP8 training stability supports the choice of FP4 FLOPs fraction as an efficiency proxy. Or just remove it?} 
Therefore, we approximate $E$ as the fraction of FLOPs executed in FP4 format, where the remaining $1-E$ fraction of FLOPs is executed using the FP8 format. This provides a practical proxy for efficiency in our evaluation.

\subsection{Integer Linear Programming Problem}
\label{sec:ILP}
% Efficiently balancing training quality and computational efficiency for LLMs is a non-trivial challenge due to the diverse behaviors of different layers during quantization. Traditional heuristics, such as manually assigning higher precision to sensitive layers, fail to systematically consider the global impact of both quality and efficiency.

The problem of finding the optimal quantization settings that meet the efficiency improvement requirements while minimizing the training quality can be mapped to a modified knapsack problem. 
Therefore, we could solve the problem by a Integer Linear Programming Problem (ILP), which ensures globally optimal solutions under the given constraints.

In our case, the \textit{value} corresponds to the negative of training quality loss 
$Q$, and the \textit{weight} corresponds to the negative of efficiency improvement 
$E$. Each layer of the model is viewed as an "item", with multiple possible quantization settings from which only one can be selected. For each layer $i$, the options are combinations of FP8 and FP4 formats for inputs, weights, and gradients. With $m$ layers and $n$ options per layer, our objective is to minimize overall quality loss while achieving the targeted efficiency improvement $E_t$.

We define the following symbols for clarity:
\begin{itemize}
\item $q_{i,j}$ is the quality loss of selecting the \(j\)-th option for layer \(i\).
\item $e_{i,j}$ is the efficiency saving of selecting the \(j\)-th option for layer \(i\).
%\item $\mathcal{X}$ is the set of all valid binary decision variables across layers and options, representing the feasible space for selection.
\item $x_{i,j} \in  \{0, 1\}$ is a binary decision variable indicating if the \(j\)-th option for layer \(i\) is selected.
\item $E_t$ is a floating point number in $[0, 1]$ that represents the target efficiency improvement. For example, $E_t=0.5$ indicates that half of the FLOPs are executed in FP4.
\end{itemize}

The ILP formulation guarantees an optimal solution exists for any $E_t$ in $[0,1]$. When $E_t=0$, the optimal solution assigns FP8 precision to all linear layers. And when $E_t=1$, the optimal solution assigns FP4 precision to all linear layers.

\begin{align}
\left\{x_{i,j}^*\right\} = \arg\min_{\{x_{i,j}\}} & \sum_{i=0}^{m-1} \sum_{j=0}^{n-1} q_{i,j} x_{i,j}\\
\text {s.t.} &\sum_{i=0}^{m-1} \sum_{j=0}^{n-1} e_{i,j} x_{i,j} \geq E_t \\
    &\sum_{j=0}^{n-1} x_{i,j} = 1, \quad \forall i \in \{0, 1, \dots, m-1\} \\
    &x_{i,j} \in \{0, 1\}, \quad \forall i,j
\end{align}

%\yyy{what does $x_{i,j} \in \{0, 1\}, \forall x_{i,j} \in \mathcal{X}$ mean here ? Do you mean $ \{x_{i,j}\} \in \mathcal X$?}

\subsection{Incorporating Pipeline Parallelism}
Pipeline parallelism~\cite{narayanan2021efficient} is widely used in LLM training to enhance memory and compute efficiency by partitioning the model's layers into stages that process data concurrently. However, imbalances in computation time across pipeline stages can create bubbles, which bottleneck efficiency improvements at the slowest stage.
To address this, we extend the ILP problem to explicitly consider pipeline parallelism by ensuring balanced efficiency savings across all stages. We model this as a grouped knapsack problem, where each group represents a pipeline stage, and the objective is to select an optimal quantization scheme per layer while maintaining balanced efficiency across stages.

We set some additional definitions.
$K$ is the number of groups, and
$g$ is the number of items in every group.
% $T$ is the minimum weight of items in every group.
% $W_k = \sum_{i=k*g}^{(k+1)*g-1} \sum_{j=0}^{n-1} w_{i,j} x_{i,j} $ is the total weights of items in $k$-th group.
Formally, we modify the prior ILP formulation by replacing the efficiency constraint line (2) with the following group-aware (pipeline-stage-aware) constraint:

\begin{align}
    \text{s.t. } & \sum_{i=k*g}^{(k+1)*g-1} \sum_{j=0}^{n-1} e_{i,j} x_{i,j} \geq \frac{E_t}{K}, \forall k \in {0, 1, \cdots K}
\end{align}

% \begin{align*}
% \text {minimize } & \sum_{i=0}^{m-1} \sum_{j=0}^{n-1} q_{i,j} x_{i,j} \\
% \text{s.t. } & \sum_{i=k*g}^{(k+1)*g-1} \sum_{j=0}^{n-1} e_{i,j} x_{i,j} \geq E_t / K, \forall k \in {0, 1, \cdots K} \\
% &\sum_{j=0}^{n-1} x_{i,j} = 1, \quad \forall i \in \{0, 1, \dots, m-1\} \\
% &x_{i,j} \in \{0, 1\}, \quad \forall x_{i,j} \in \mathcal{}{X}
% \end{align*}

This ensures that each pipeline stage contributes equally to the overall efficiency target, preventing bottlenecks and maintaining a well-balanced workload across stages.

% \subsection{Customization of Efficiency and Quality Metrics}
% For efficiency improvement $E$, while our current formulation approximates $E$ as the fraction of FLOPs executed in FP4 format due to the lack of access to hardware that supports FP4 and FP8 GEMM operations. However, a more accurate representation can be achieved by incorporating the actual runtime of FP8 and FP4 GEMMs or ans accurate model of their runtime. Such a model would consider key hardware properties, including memory bandwidth, compute intensity, and GPU architecture. By integrating these runtime models, $E$ can more accurately reflect practical efficiency gains, enabling the generation of quantization schemes optimized for end-to-end runtime and energy improvements.

% Additionally, the flexibility of our framework allows for straightforward modifications to the quality loss metric $Q$ to create different quantization schemes tailored to specific objectives. For example, To generate the quantization schemes that minimize the absolution quantization error per tensor, we could modify the quality loss $Q$ to be $=\sum \FrobNorm{x-q(x)}$ where $x$ represents the inputs, weights and gradients for every linear layers, and $q(x)$ denotes the corresponding quantized tensors, to generate the quantization schemes that minimize the relative quantization error per tensor, we could modify the quality loss $Q$ to be $=\sum \FrobNorm{(x-q(x))/x}$.

% This flexibility broadens the applicability of our approach across diverse quantization strategies and hardware.
\section{Evaluation}
\label{sec:evaluation}
%% Put table here
\begin{table*}[t]
\centering
\scriptsize
\caption{Accuracy comparison of test benchmarks across different quantization schemes for TinyLlama at 50k checkpoint. We \textbf{boldface} the best performing quantization schemes with the fixed efficiency (fractions of FP4 FLOPS) from 25\% to 75\%. It demonstrates that \THISWORK\ consistently outperforms all the other quantization schemes, and is very close the the BF16 baseline.}
\label{tab:tinyllama-50k-all}
\begin{tabular}{c|c|c|c|c|c|c|c|c|c|c}
\hline
\multicolumn{1}{c|}{\multirow{2}{*}{\shortstack{{Fraction of} \\ {FP4 FLOPS}}}} & \multirow{2}{*}{Quant schemes} & \multicolumn{2}{c|}{Math} & \multicolumn{1}{c|}{Aggregate} & \multicolumn{5}{c|}{Commonsense} & \multicolumn{1}{c}{\multirow{2}{*}{Average}} \\ \cline{3-10}
\multicolumn{1}{c|}{} &  & \multicolumn{1}{c|}{ARC\_c} & \multicolumn{1}{c|}{ARC\_e} & \multicolumn{1}{c|}{MMLU} & \multicolumn{1}{c|}{BoolQ} & \multicolumn{1}{c|}{HellaSwag} & \multicolumn{1}{c|}{Obqa} & \multicolumn{1}{c|}{PiQa} & \multicolumn{1}{c|}{WinoGrande} & \multicolumn{1}{c}{} \\ \hline\hline
\multirow{2}{*}{0} & BF16 & 24.15 & 46.09 & 26.56 & 59.72 & 43.94 & 31.00 & 67.41 & 54.85 & 44.22 \\
 & FP8 & 23.98 & 45.71 & 26.61 & 60.15 & 44.04 & 31.20 & 67.46 & 54.46 & 44.20\\ \hline
\multirow{7}{*}{25\%} & \THISWORK & 24.06 & 45.75 & 26.73 & 60.28 & 43.84 & 31.20 & 67.25 & 54.78 & \textbf{44.24} \\
 & min-abs-err & 24.06 & 45.75 & 26.72 & 60.55 & 43.99 & 30.20 & 67.57 & 53.99 & 44.10 \\
 & min-rel-err & 24.49 & 45.71 & 26.22 & 60.52 & 43.72 & 30.60 & 67.68 & 54.30 & 44.16 \\
 & random0 & 23.21 & 39.44 & 25.32 & 59.42 & 35.58 & 28.40 & 63.76 & 51.54 & 40.83 \\
 & random1 & 23.98 & 29.08 & 25.69 & 39.39 & 28.22 & 23.80 & 54.08 & 49.49 & 34.22 \\
 & random2 & 24.06 & 44.65 & 26.14 & 61.25 & 43.07 & 30.00 & 67.30 & 53.75 & 43.78 \\
 & E-layer-type & 27.82 & 26.43 & 22.95 & 37.83 & 26.01 & 25.60 & 48.97 & 50.04 & 33.21 \\
\hline
\multirow{8}{*}{50\%} & \THISWORK & 24.06 & 45.41 & 27.04 & 60.43 & 43.79 & 30.40 & 67.57 & 54.54 & \textbf{44.16} \\
 & min-abs-err & 25.60 & 26.98 & 25.41 & 38.04 & 26.50 & 25.80 & 51.25 & 50.04 & 33.70 \\
 & min-rel-err & 27.56 & 28.16 & 24.39 & 37.83 & 26.64 & 26.40 & 52.34 & 51.22 & 34.32 \\
 & random0 & 26.62 & 27.65 & 23.09 & 37.83 & 27.26 & 24.60 & 51.96 & 49.64 & 33.58 \\
 & random1 & 28.41 & 27.53 & 24.79 & 39.36 & 27.88 & 26.20 & 51.63 & 49.57 & 34.42 \\
 & random2 & 24.15 & 44.61 & 25.76 & 61.25 & 42.74 & 30.40 & 66.92 & 53.04 & 43.61 \\
 & E-layer-id & 26.37 & 26.56 & 24.80 & 37.83 & 27.19 & 25.20 & 49.46 & 48.70 & 33.26\\
 & E-layer-type & 26.71 & 26.98 & 22.94 & 37.83 & 26.47 & 24.20 & 50.22 & 49.57 & 33.11\\ \hline
\multirow{6}{*}{75\%} & \THISWORK & 24.40 & 46.04 & 26.68 & 60.76 & 43.46 & 30.80 & 67.41 & 54.14 & \textbf{44.21} \\
 & min-abs-err & 27.22 & 26.52 & 24.66 & 37.83 & 26.40 & 25.40 & 50.92 & 49.25 & 33.53 \\
 & min-rel-err & 25.94 & 26.18 & 24.28 & 37.83 & 26.50 & 25.80 & 51.09 & 50.67 & 33.54 \\
 & random0 & 27.56 & 27.15 & 25.30 & 37.83 & 26.21 & 26.20 & 51.14 & 49.88 & 33.91 \\
 & random1 & 28.67 & 27.57 & 24.75 & 37.83 & 25.99 & 24.60 & 51.41 & 50.20 & 33.88 \\
 & random2 & 27.39 & 27.36 & 25.74 & 37.83 & 26.56 & 26.00 & 50.76 & 49.33 & 33.87 \\ \hline
80\% & \THISWORK & 24.23 & 46.00 & 26.83 & 61.07 & 43.16 & 31.80 & 67.36 & 53.67 & \textbf{44.27} \\ \hline
85\% & \THISWORK & 25.60 & 26.64 & 26.50 & 37.83 & 26.55 & 25.80 & 52.99 & 50.20 & 34.01 \\ \hline
100\% & FP4 & 26.11 & 27.02 & 23.12 & 37.83 & 25.97 & 23.60 & 50.33 & 51.46 & 33.18
\end{tabular}
\end{table*}

\subsection{Experiment Setup}
\noindent\textbf{Models.}
We conduct our pretraining experiments using open-source models and datasets, leveraging publicly available intermediate checkpoints to ensure reproducibility and accessibility. Specifically, we use the TinyLlama 1B~\cite{zhang2024tinyllama} and OpenLlama 3B, OpenLlama 7B~\cite{openlm2023openllama} models, as well as an industry LLaMA-style dense 70B model. Given that training TinyLlama 1B from scratch on 300B tokens requires approximately 3,456 A100 GPU hours~\cite{zhang2024tinyllama}, we opt to resume pretraining from the released intermediate checkpoints for open source models. 

Specifically, for the TinyLlama 1B model, we use checkpoints at 5K steps (10B tokens), 10K steps (21B tokens), 20K steps (42B tokens), 50K steps (105B tokens), and 240K steps (503B tokens). For OpenLlama 3B and 7B models, we use checkpoints at 50k steps and 100k steps. Since the released checkpoints lack optimizer states, which are crucial for estimating weight divergence, we resume pretraining for a few more steps using Huggingface. We resume training from these checkpoints to evaluate different quantization schemes.

For the 70B model, we pretrain in BF16 precision for 10K steps and continue training up to 25K steps. Even excluding activations, training a 70B model requires approximately 1120 GB of GPU memory solely for model weights, gradients, and optimizer states~\cite{rajbhandari2020zero}. Storing intermediate activations during backpropagation imposes additional memory overhead. Due to the high resource demands and lack of public intermediate checkpoints for 70B models, we limit our experiments on this model to this training window only.

\noindent\textbf{Training Datasets.}
We adhere to the recommended guidelines for each model to download and preprocess the corresponding training datasets. For TinyLlama, we utilize a mixture of the SlimPajama and StarcoderData datasets, while for OpenLlama, we use the RedPajama~\cite{together2023redpajama} dataset.  Since our experiments involve resuming pretraining from intermediate checkpoints rather than training models from scratch, we sample approximately 1\% of the original datasets without hurting the training performance.
For 70B model, we use internal industry datasets.

\noindent\textbf{Quantization Format and Implementation.}
Since we do not have access to GPUs that natively support FP8 or FP4 formats, we implement fake quantization to emulate the quantize and dequantize operations during training. Our fake quantization function is designed to support a variety of floating-point formats less than 16 bits. %However, in our experiments, we focus on the FP8\_E3M4 and FP4\_E2M1 formats, \hanmei{I don't get it, how about other formats?} as these have been empirically shown to exhibit the lowest quantization error among formats with the same bitwidth.
Following the DeepSeek-V3~\cite{liu2024deepseek} FP8 training recipe, we apply 1x128 tile-wise quantization for input activations and gradients, and 128x128 block-wise quantization for weights. This configuration is chosen to balance precision and computational efficiency while minimizing quantization-induced errors. 
When applying the FP4 format to output gradients, we utilize stochastic rounding~\cite{croci2022stochastic} that avoids training to stagnate by probabilistically rounding the values to the two nearest values~\cite{courbariaux2016binarized, li2017training,chmiel2023accurate}.

\noindent\textbf{Baselines}
To our knowledge, no prior work systematically searches for fine-grained layer-specific quantization schemes for LLM pre-training. To evaluate the effectiveness of \THISWORK, we compare it against the following baselines:
\begin{itemize}
    \item Uniform Precision: Models trained with uniform precision across all layers using BF16, FP8 or FP4.
    \item Empirical Coarse-grained Quantizations: Higher precision is assigned to sensitive layers based on the empirical observations in prior work. This includes the first and last few layers (\textit{E-layer-id})~\cite{dumitru2024layer,zhang2024investigating} and MLP layers in each transformer block (\textit{E-layer-type}).
    \item Fine-Grained Quantization by Error Minimization: Layer-specific configurations optimized to minimize absolute (\textit{min-abs-err}) or relative quantization error (\textit{min-rel-err}) for each tensor, focusing on local metrics without considering overall training quality. For a fair comparison, we also use the ILP solver as in Section~\ref{sec:ILP} where the quality loss $Q$ is the absolute or relative quantization error.
    \item Random Fine-grained Quantization schemes: Precision levels are randomly assigned to each layer (\textit{random}) to evaluate robustness.
\end{itemize}

\noindent\textbf{Efficiency Metric: Fraction of FP4 FLOPs.}
Lacking access to hardware that supports both FP8 and FP4(e.g., Blackwell) for end-to-end runtime measurements, we use the fraction of FP4 FLOPs as an efficiency metric. This metric represents the proportion of FLOPs in the model's \textit{linear layers} executed with FP4 precision, while the remaining FLOPs are executed with FP8. To our best knowledge, the latest FP4 training studies ~\cite{chmiel2025fp4,tseng2025training,yang2025empirical} rely on simulation. Our goal is to guide early-stage design and estimate system-level impact ahead of broad hardware availability.

\noindent\textbf{Evaluation.}
We evaluate the LLM performance using the training loss and LM-Evaluation-Harness framework~\cite{eval-harness}, a widely adopted tool for assessing large language models (LLMs). We use training loss as one metric due to its strong correlation with quality under fixed settings. Following common practices for LLM evaluation, we cover several key categories: 1) Math and reasoning: ARC Easy (ARC-e), ARC Challenge (ARC-c)~\cite{clark2018think}; 2) Aggregate: MMLU~\cite{hendrycks2020measuring}; 3) Commonsense understanding: BoolQ~\cite{clark2019boolq}, PiQA~\cite{bisk2020piqa}, HellaSwag~\cite{zellers2019hellaswag}, OpenBookQA~\cite{mihaylov2018can} and WinoGrande~\cite{sakaguchi2021winogrande}. All evaluations are conducted in 0-shot, with the exception of MMLU, which is evaluated using 5-shot prompting.

\noindent\textbf{Software and Hardware Stack.}
Our experiments are conducted using the Huggingface training framework~\cite{wolf2019huggingface} with Distributed Data Parallelism (DDP) for efficient training. The experiments for 1B and 3B models are executed on a single node equipped with 4 NVIDIA A40 GPUs, each with 40GB of memory, while the experiments for 7B model are executed on a single node with 8 NVIDIA A100 GPUs, each with 80 GB of memory.
For 3B and 7B models, we additionally use Deepspeed~\cite{rajbhandari2020zero} Zero stage 1 that partitions the optimizer states across GPU ranks to fit the model into the GPUs.
We conduct the 70B model experiments on 64 H100 GPUs using a Megatron-like~\cite{shoeybi2019megatron} training framework, configured with fully sharded data parallelism (FSDP) = 2, tensor parallelism (TP) = 4, and pipeline parallelism (PP) = 8. %This configuration balances memory efficiency, model parallelism, and interconnect bandwidth utilization, enabling stable large-scale training while keeping per-GPU memory usage within hardware limits. 

For the solution of ILP, we utilize the \verb|scipy.optimize.milp| function, which is a wrapper of the HiGHS linear optimization software~\cite{hall2023highs}. We set the time limit for every ILP solution to 30 seconds, where it usually takes a few seconds. %The solver~\cite{hall2023highs} guarantees one optimal solution if multiple exist.

\subsection{Results}

% \begin{figure}[t]
% \centering
% \includegraphics[width=\linewidth]{figs/heatmap_tinyllama_50k_0.25.pdf} \\ 
%  \vspace{-5mm}
% \subfloat[\THISWORK]{\hspace{.33\linewidth}}
% \subfloat[min-abs-err]{\hspace{.33\linewidth}}
% \subfloat[min-rel-err]{\hspace{.33\linewidth}}
% \vspace{-3mm}
% \caption{Per-layer precision assignments at 25\% FP4 FLOPs across different quantization schemes.}
% \label{fig:heatmap_tiny_0.25}
% \end{figure}

% \begin{figure}[t]
% \centering
% \includegraphics[width=\linewidth]{figs/heatmap_tinyllama_50k_0.5.pdf}\\ 
%  \vspace{-5mm}
% \subfloat[\THISWORK]{\hspace{.33\linewidth}}
% \subfloat[min-abs-err]{\hspace{.33\linewidth}}
% \subfloat[min-rel-err]{\hspace{.33\linewidth}}
% \vspace{-3mm}
% \caption{Per-layer precision assignments at 50\% FP4 FLOPs across different quantization schemes.}
% \label{fig:heatmap_tiny_0.5}
% \end{figure}
\begin{figure}
    \centering
    \includegraphics[width=\linewidth]{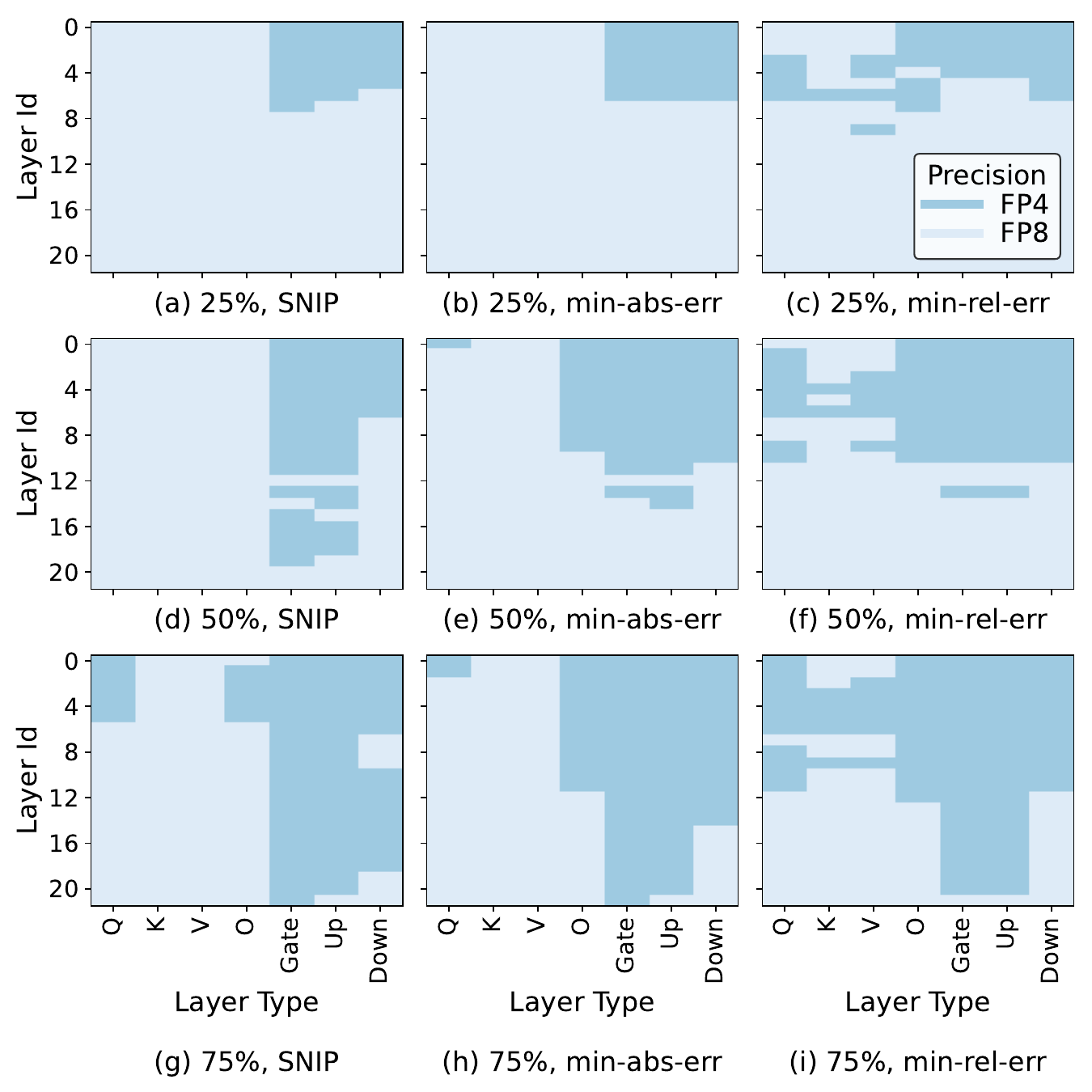}
    \caption{Per-layer precision assignments at 25\%, 50\%, and 75\% FP4 FLOPs across different quantization schemes.}
    \label{fig:heatmap_tiny_all}
\end{figure}

\begin{figure}[t]
    \centering
    \includegraphics[width=0.9\linewidth]{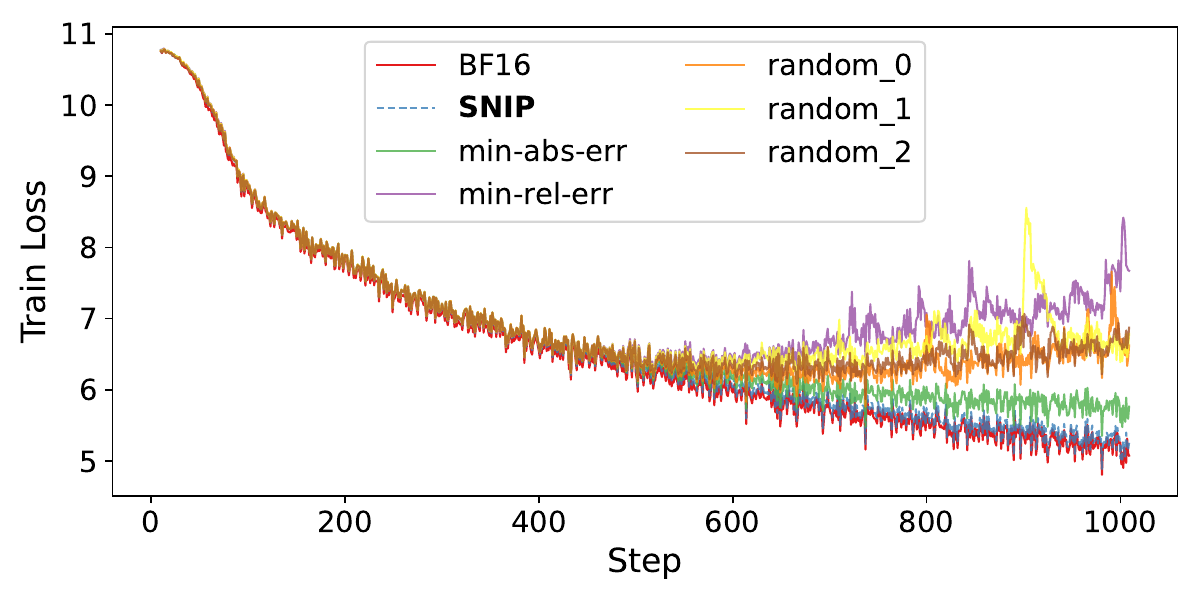}
    \caption{Training loss curve for BF16 (baseline) and different quantization schemes under a 75\% FP4 FLOPs efficiency budget for TinyLlama 1B model.}
    \label{fig:train_loss_tinyllama_0k}
\end{figure}

\begin{figure}[t]
    \centering
    \includegraphics[width=\linewidth]{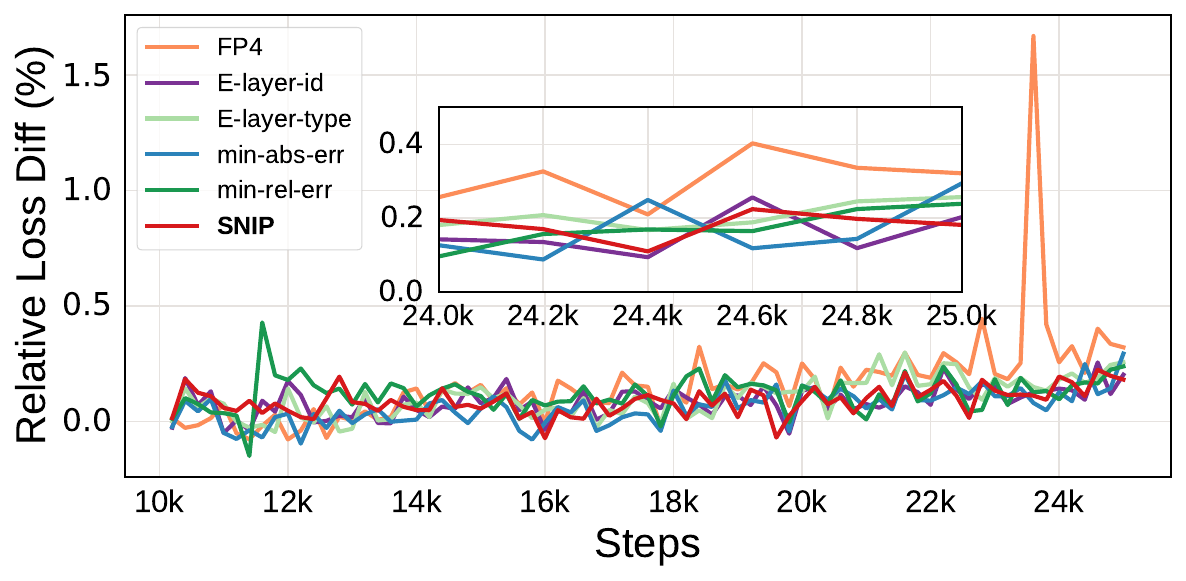}
    \caption{Relative training loss difference over BF16 (baseline) for the Llama 70B dense model from 10k to 25k steps. \textit{FP4} means all layers are using FP4 precision. And different quantization schemes are under a 50\% FP4 FLOPs efficiency budget.}
    \label{fig:70b_loss_diff}
\end{figure}

\subsubsection{\THISWORK\ Outperforms Other Quantization Schemes}

In this section, we use extensive experiment results on different checkpoints, different models, and different efficiency savings to demonstrate the effectiveness of the \THISWORK\ over all the other baselines. 

%\hanmei{training from 51k to 52k steps: does this mean you only train 1k steps from checkpoint at 51k? If this is the case, people will argue that 1k is not enough. Plus, you mentioned that ``updating every 50k to 100k steps is sufficient'', how many quantization scheme updates are done in 1k steps?}

\noindent\textbf{Across Different Efficiency Savings}
Table~\ref{tab:tinyllama-50k-all} highlights the consistent superiority of \THISWORK\ over other quantization schemes for the TinyLlama model at the 50k-step checkpoint, across all test benchmarks and FP4 FLOP fractions, particularly as the required efficiency savings (fractions of FP4 FLOPs) increase. For fractions of 25\%, 50\%, and 75\% FP4 FLOPs, \THISWORK\ achieves the highest average scores, closely approximating the BF16 baseline (44.22), while all other methods experience a significant decline in performance. For instance, at 25\% FP4 FLOPs, \THISWORK\ achieves an average score of 44.24, outperforming all other quantization schemes. Notably, some methods, such as \textit{min-abs-err} and \textit{min-rel-err}, also maintain relatively high accuracy, though they still fall short of \THISWORK. At 50\% FP4 FLOPs, \THISWORK\ maintains its high performance with an average score of 44.16, while methods like \textit{min-abs-err} and \textit{min-rel-err} fail to retain competitive results. %\hanmei{When reading this paragraph, I'm curious about why min error fails? Intuitively, they are expected to perform well. Later I found answers in the next next paragraph. I think you should move it here.} 
The divergence becomes more pronounced at 75\% FP4 FLOPs, where \THISWORK\ achieves an average score of 44.21, significantly outperforming random baselines and heuristic methods, which is close to the BF16 baseline. We explain the reasons for the poor performance of other quantization schemes in the following paragraphs.

Coarse-grained quantization schemes, such as \textit{E-layer-type} and \textit{E-layer-id}, leverage empirical knowledge to assign FP8 precision to specific layer types (e.g., MLPs or down-projection layers) or layer indices (e.g., the first and last layers) deemed more sensitive to precision loss, while using FP4 for the remaining layers. While effective at lower efficiency budgets (e.g., 25\% FP4 FLOPs) by conservatively prioritizing sensitive layers, they lack the adaptability to fully utilize the fine-grained design space of per-layer quantization. At higher efficiency budgets (e.g., 50\% FP4 FLOPs), these schemes demonstrate robustness compared to other quantizations but still fall short of the superior performance achieved by \THISWORK, which consistently delivers better accuracy across all efficiency budgets.

Layer-wise heuristics, such as \textit{min-abs-err} and \textit{min-rel-err}, prioritize minimizing local quantization errors while ignoring global training dynamics. This approach works well at lower FP4 fractions (e.g., 25\%), but as the FP4 fraction increases, the lack of global information leads to significant performance degradation.

Figures~\ref{fig:heatmap_tiny_all} visualize the per-layer precision assignments of different quantization schemes with different efficiency budgets. At a lower efficiency budget, like 25\% FP4 FLOPs, \THISWORK\ makes layer-wise precision choices similar to those of \textit{min-abs-err} and \textit{min-rel-err}, resulting in comparable accuracy, as shown in Table~\ref{tab:tinyllama-50k-all}. However, at 50\% FP4 FLOPs, clear differences emerge; \textit{min-abs-err} and \textit{min-rel-err} typically assign lower precision to early layers with minimal quantization error, potentially ignoring their aggregate effect on training loss. At a 75\% FP4 FLOPs efficiency, \THISWORK\ opts for higher precision in the down projection layers of middle layers, unlike \textit{min-rel-err}, which focuses on the final layers.
\THISWORK\ demonstrates a more balanced and globally optimized precision assignment compared to the other two methods.

%This highlights the necessity of globally informed strategies, like \THISWORK, which take overall training quality into account. The results further emphasize the robustness of \THISWORK\ in maintaining model quality under aggressive efficiency constraints, where other methods fail to deliver consistent performance.

\noindent\textbf{Training Loss when Training From Scratch} 
Figure~\ref{fig:train_loss_tinyllama_0k} presents the training loss over 1k steps for the TinyLlama model, trained from scratch with BF16 (baseline) and various quantization schemes under a 75\% FP4 FLOPs efficiency budget, all using the same hyperparameters. The loss curves for BF16 and \THISWORK\ nearly overlap, with \THISWORK\ exhibiting a slightly higher training loss than BF16. Specifically, the training loss is 5.27 for BF16 and 5.34 for \THISWORK. In contrast, all other quantization schemes with the same efficiency budget fail to maintain stable training and exhibit clear divergence, underscoring the effectiveness of \THISWORK\ in preserving training stability while achieving substantial efficiency gains.

\begin{table}[t]
\centering
\scriptsize
\caption{Accuracy comparison across quantization schemes under a fixed efficiency budget, evaluated at different checkpoints and model sizes. BF16 serves as the baseline.}
\label{tab:all_0.75}
\begin{tabular}{c|c|c|c|c|c|c|c}
\hline
Model & \multicolumn{3}{c|}{TinyLlama 1B} & \multicolumn{2}{c|}{OpenLlama 3B} & \multicolumn{2}{c}{OpenLlama 7B} \\ \hline
Checkpoint & 5k & 50k & 240k & 50k & 100k & 50k & 100k \\ \hline \hline
BF16 & 39.16 & 44.22 & 46.13 & 49.94 & 51.85 & 52.46& 54.42\\ \hline
\THISWORK & \textbf{39.15} & \textbf{44.21} & \textbf{45.78} & 49.72 & 51.72 & \textbf{52.43} & \textbf{54.45}\\ \hline
min-abs-err & 39.07 & 33.53 & 45.54 & 49.95 & \textbf{51.99} & 52.15 & 54.40 \\ \hline
min-rel-err & 33.11 & 33.54 & 45.57 & \textbf{50.19} & 51.93 & 52.42 & 54.34\\ \hline
random0 & 33.71 & 33.91 & 45.23 & 48.16 & 49.59 & 49.52 & 52.87\\ \hline
random1 & 33.24 & 33.88 & 32.86 & 49.16 &  51.04 & 49.05 & 51.84\\ \hline
random2 & 38.96 & 33.87 & 33.55 & 48.35 & 51.32 & 50.13 & 52.08\\ \hline
\end{tabular}
\end{table}
\begin{table}[t]
\centering
\scriptsize
\caption{Accuracy comparison over BF16 across different quantization schemes under a 50\% efficiency budget for the Llama 70B dense model. }%Baseline accuracies using FP8 and FP4 precision is also provided for reference.}
\label{tab:70b_eval}
\begin{tabular}{c|c|c|c}
\hline
Model & ARC\_c & MMLU & HellaSwag\\ \hline \hline
FP8 & -0.86 & -0.01 & 0.24 \\ \hline
FP4 & 0.17 & -0.37 & 0.18 \\ \hline
\THISWORK & -0.17 & -0.12 & 0.59 \\ \hline
E-layer-id & -0.52 & -0.73 & -0.04 \\ \hline
E-layer-type & -0.43 & 0.01 & -0.07\\ \hline
min-abs-err & 0.94 & -0.57 & 0.27 \\ \hline
min-rel-err & -0.77 & -0.45 & 0.09 \\ \hline
\end{tabular}
\end{table}
\noindent\textbf{Across Different Checkpoints and Models}
Table~\ref{tab:all_0.75} highlights the robustness of \THISWORK\ across different models and training phases. The efficiency budget is 75\% for TinyLlama 1B model and 50\% for OpenLlama 3B, 7B model because OpenLlama models are more sensitive to precision loss. It consistently delivers competitive accuracy for TinyLlama 1B and OpenLlama 3B and 7B, demonstrating scalability to larger models. Additionally, \THISWORK\ maintains stable performance across various training checkpoints, ensuring reliability throughout different training stages. While in OpenLlama 3B, \THISWORK\ shows slightly lower test accuracy than \textit{min-abs-err} and \textit{min-rel-err}, its training loss remains consistently lower. The fact that \textit{min-abs-err} and \textit{min-rel-err} outperform the BF16 baseline in test accuracy suggests potential noise in their evaluations rather than true improvements. 

We further evaluate the training stability of quantization on the 70B dense model by measuring the relative training loss difference with respect to BF16 precision from 10k to 25k steps (Figure~\ref{fig:70b_loss_diff}). A lower value indicates better stability. When all layers are trained with FP4 precision, the model exhibits a gradual divergence, though notably slower than that observed in smaller models such as TinyLlama-1B (Figure~\ref{fig:train_loss_tinyllama_0k}). This observation is consistent with prior results on quantization for LLM inference~\cite{frantar2022gptq,malinovskii2024pv,li2025icquant,chen2025scaling}, which show that larger models are more resilient to precision loss. Figure~\ref{fig:70b_loss_diff} further shows that, under a 50\% FP4 FLOPs efficiency budget, our \THISWORK\ scheme achieves loss curves that remain closely aligned with BF16, while heuristic schemes, such as \textit{min-rel-err} and \textit{E-layer-type} (assigning FP8 to the gate and up-projection layers of the MLP while using FP4 for the remaining layers), exhibit occasional spikes and larger deviations. It can be seen from the figure that \THISWORK\ and \textit{E-layer-id} (assigning FP4 to the middle 50\% of layers and FP8 to the rest) outperforms other quantization schemes.

Table~\ref{tab:70b_eval} reports accuracy differences over BF16 on evaluation datasets. "+" means improvement and "-" means degradation. FP4 baselines suffer from accuracy drops in MMLU, while heuristic selection schemes produce inconsistent results across tasks. In contrast, SNIP consistently delivers stable accuracy, and in some cases even surpasses FP4 and FP8 baselines. Taken together, these results show that SNIP not only provides smoother training dynamics but also achieves superior downstream accuracy compared to heuristic quantization strategies across models from 1B, 3B, 7B to 70B.

%These results highlight the adaptability and reliability of \THISWORK\ across diverse training models and stages.

\subsection{In-Depth Analysis}
\begin{figure}
    \centering
    \includegraphics[width=0.95\linewidth]{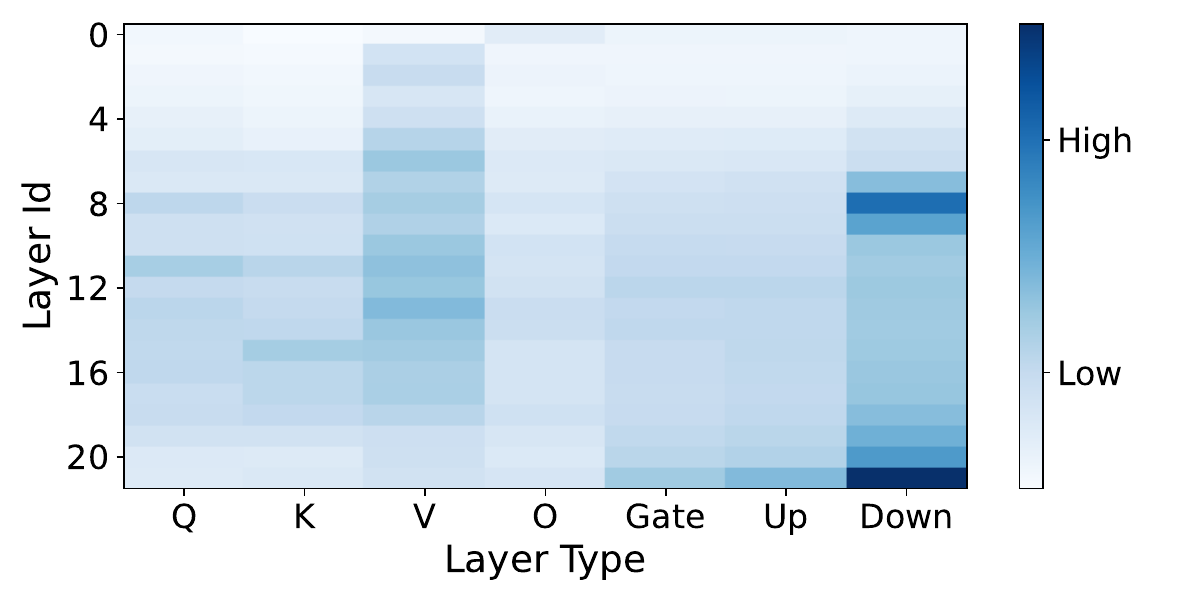}
    \caption{Heatmap of the layer-wise quality loss to FP4 quantization for 1B model at 50k step checkpoint. The darker regions indicate higher sensitivity to FP4 precision loss. }
    \label{fig:layer_importance}
\end{figure}
\noindent\textbf{Quality Importance of Each Layer.}
%In Section~\ref{sec:optimization}, we present the formulation of the optimal per-layer quantization policy, balancing quality loss ($Q$) and efficiency savings ($E$). 
To better understand the impact of quantizing each layer in FP4 format, we visualize the “importance” of each layer in Figure~\ref{fig:layer_importance}, which corresponds to quality loss (Q) in Section~\ref{sec:optimization}. % This importance inherently captures both the quantization error of each tensor and its sensitivity to precision loss, ensuring that our quality loss metric is practical and accurate.
The figure highlights key observations: the last layer's MLP stands out as the most critical, requiring higher precision to maintain model quality. Similarly, down-projection (Down) layers in the MLP structure, particularly in the later layers, show higher sensitivity, emphasizing their importance in preserving transformed information. Among the attention layers (Q, K, V), the V (Value) layers stand out as more sensitive than Q (Query) and K (Key) layers, despite operating on the same input activations, emphasizing their importance in the attention mechanism.

These findings underscore the need for fine-grained, layer-specific quantization strategies, as demonstrated by \THISWORK, to effectively balance quality and efficiency while providing insights into model behavior.

\begin{figure}
\centering
\includegraphics[width=\linewidth]{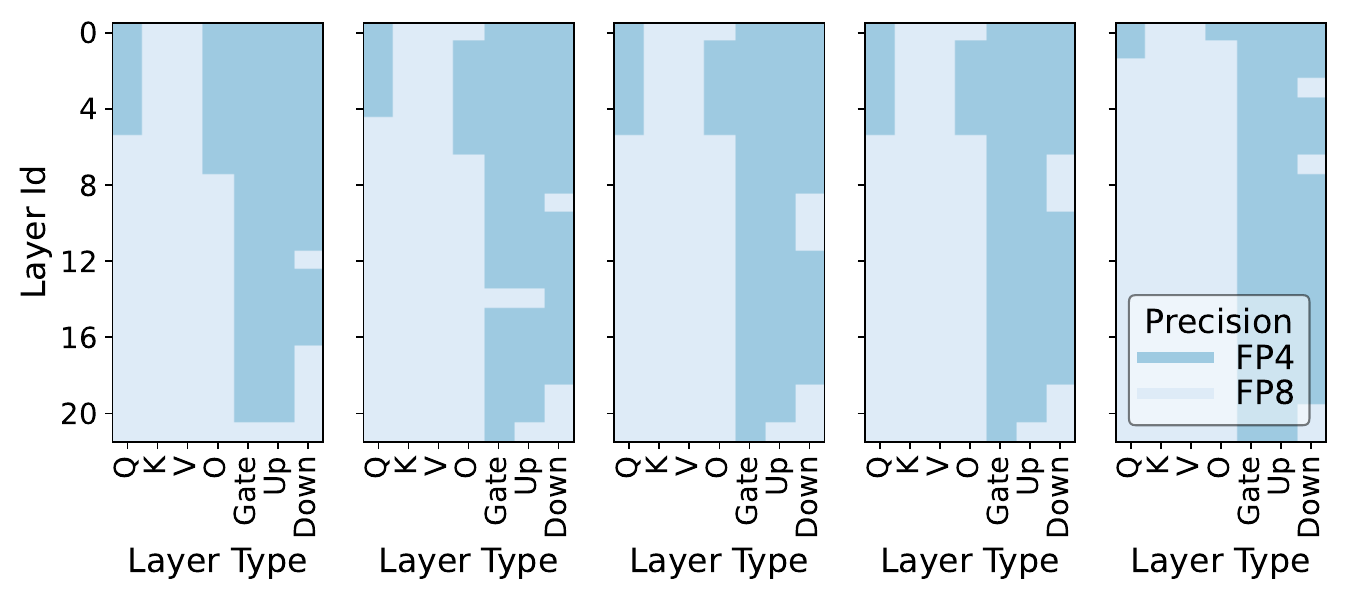} \\ 
 % \vspace{-5mm}
% \subfloat[5k]{\hspace{.2\linewidth}}
% \subfloat[10k]{\hspace{.2\linewidth}}
% \subfloat[20k]{\hspace{.2\linewidth}}
% \subfloat[50k]{\hspace{.2\linewidth}}
% \subfloat[240k]{\hspace{.2\linewidth}}
\noindent\begin{minipage}{\linewidth}
  \centering
  \begin{minipage}{0.18\linewidth}\centering {\small (a) 5k}\end{minipage}\hfill
  \begin{minipage}{0.18\linewidth}\centering {\small (b) 10k}\end{minipage}\hfill
  \begin{minipage}{0.18\linewidth}\centering {\small (c) 20k}\end{minipage}\hfill
  \begin{minipage}{0.18\linewidth}\centering {\small (d) 50k}\end{minipage}\hfill
  \begin{minipage}{0.18\linewidth}\centering {\small (e) 240k}\end{minipage}
\end{minipage}

\caption{Evolution of per-layer precision assignments determined by \THISWORK\ at 75\% FP4 FLOPs across different training checkpoints (5k, 10k, 20k, 50k, 240k) for TinyLlama model.}
\label{fig:heatmap_tinyllama_0.75}
\end{figure}
\noindent\textbf{Generate Quantization Scheme Periodically.}
Figure~\ref{fig:heatmap_tinyllama_0.75} demonstrates how \THISWORK\ adjusts per-layer precision assignments across training checkpoints, reflecting the evolving importance of layers. From 5k to 50k steps, precision assignments remain stable, suggesting little change in layer sensitivities in a short time period. At 240k steps, clear differences emerge, with more layers assigned higher precision (FP8). Notably, the first few layers gain importance in later stages, receiving higher precision, while the last few layers shift to lower precision (FP4). This highlights \THISWORK's ability to adapt precision dynamically, balancing efficiency and model quality as training progresses. Based on these findings, we recommend generating updated quantization schemes periodically, approximately every 100k steps. 
\THISWORK\ introduces runtime overheads due to the collection of statistics and processing. Steps 1 to 3 involve three additional training iterations, each incurs an overhead of roughly 2-3 times that of a normal training iteration, approximately 10 minutes. Steps 4 and 5, which handle optimization, are offloaded to the CPU, allowing GPU training to continue unaffected and take about 15 minutes. This setup ensures minimal disruption to the training process.
%\hanmei{Any runtime overhead analysis? as collecting metrics and two forward+backward incurs overhead. Just mentioning a percentage is enough.}

\begin{figure}
    \centering
    \includegraphics[width=\linewidth]{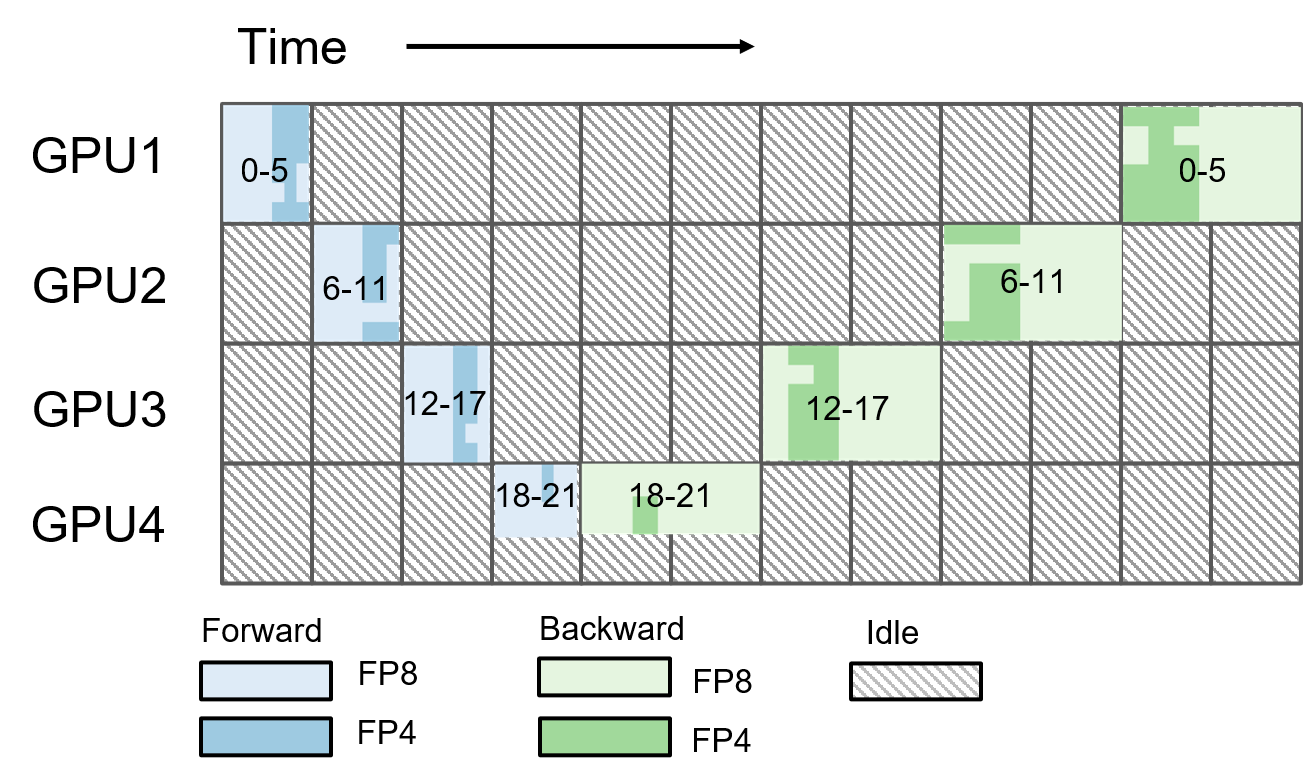}
    \caption{Timeline of pipeline parallelism of TinyLlama model using \THISWORK\ with 50\% efficiency savings with four stages. Each stage processes specific layers, marked by their layer id, and layer-specific precisions are visualized in 2D heatmaps.}%, where the x-axis represents layer type and the y-axis represents layer id.}
    \label{fig:pp_timeline}
\end{figure}
\noindent\textbf{Incorporating Pipeline Parallelism.}
Figure~\ref{fig:pp_timeline} illustrates the timeline of pipeline parallelism for the TinyLlama model using \THISWORK's fine-grained mixed precision training with 50\% efficiency savings and 4 pipeline stages for forward and backward pass. The TinyLlama model consists of 22 layers, which are evenly distributed across the first 3 stages, each processing 6 layers, while the 4th stage processes the final four layers. Layer-specific precisions, determined by \THISWORK, are visualized in 2D heatmaps for each stage, where the x-axis represents layer types, and the y-axis represents layer IDs. Note that although we have less than 50\% FP4 FLOPs in the 4th stage, since it processes 4 layers rather than 6 layers in the other 3 stages, the overall efficiency remains balanced across the pipeline. The efficient distribution of layers and mixed precision assignments helps minimize idle time and maximize overall throughput in pipeline parallelism.

\noindent\textbf{The Estimated vs. Actual Error Propagation.}
\begin{figure}
    \centering
    \includegraphics[width=\linewidth]{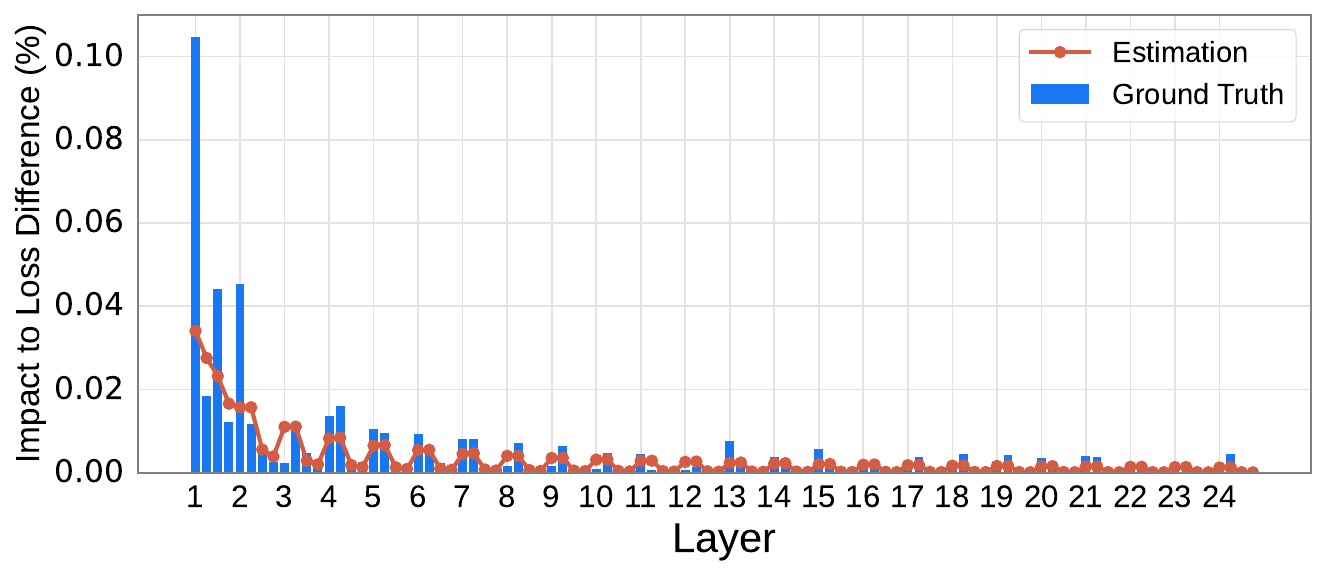}
    \caption{\THISWORK-estimated vs. ground-truth per-layer loss impact when quantizing weights and activations, showing close alignment across all layers.}
    \label{fig:loss_diff_est_gt}
\end{figure}
To ensure the reliability of our estimation method before integrating it into the ILP formulation, we validated the estimated error propagation against actual measured errors. In the experiments, we quantize each layer individually and perform a forward pass to measure the loss difference relative to the BF16 baseline, obtaining the ground-truth impact. We then apply the expression described in Section~\ref{sec:loss_divergence_fwd} to estimate the loss impact when quantizing the weights and activations of different layers. As shown in Figure~\ref{fig:loss_diff_est_gt}, the estimated per-layer impact on forward-pass loss closely matches the ground-truth measurements across all layers, capturing both the relative magnitudes and overall trends. This strong agreement demonstrates that our estimation provides an accurate proxy for actual error propagation, enabling effective layer selection in the subsequent ILP optimization.

\noindent\textbf{Memory Overhead of \THISWORK.}
%According to the overall workflow of \THISWORK\ in Figure~\ref{fig:overall}, Step~1 collects statistics during the forward and backward passes to estimate both loss divergence and weight divergence. We improve the perturbation-sensitivity estimation by replacing the global Frobenius norm with a row-wise formulation, enabling finer-grained capture of variations across input rows. For an $M \times N$ tensor, this requires storing only M or N additional values (depending on whether weights, activations, or gradients are being measured), which is negligible compared to the tensor size. In our experiments, the GPU memory overhead introduced by \THISWORK\ is less than 1\%. 
\THISWORK\ collects statistics during the forward and backward passes to estimate loss and weight divergence. To improve sensitivity estimation, we replace global Frobenius norms with a row-wise formulation, which stores only $M$ or $N$ additional values for an $M \times N$ tensor. This overhead is negligible relative to tensor size, and in practice the GPU memory overhead of \THISWORK\ is under 1\%.
% BF16: 60.35 GB, my: 59.4596 GB?

% \subsubsection{Pretraining Evaluations}

% \subsubsection{Insights of Quantization Scheme}

% Here 
% \noindent\textbf{Train Loss Can Be Misleading}

% \noindent\textbf{Same Model at Different Checkpoints.}

% \noindent\textbf{Similar Model Structure with Differnt Sizes.}
\section{Related Work}
\noindent\textbf{Mixed Precision Training for LLM Pretraining.}
Mixed precision training has emerged as a technique to balance computational efficiency and model accuracy in the training of large-scale deep learning models. It was first introduced in~\cite{micikevicius2018mixed}, which demonstrated the feasibility of training deep neural networks using FP16 for most operations, while maintaining FP32 master copies of weights to ensure numerical stability during updates. Subsequent advancements explored lower-precision formats for GEMM operations, such as FP8~\cite{liu2024deepseek, peng2023fp8,perez2023training}, FP4~\cite{wang2025optimizing}, MX format~\cite{rouhani2023microscaling,tseng2025training}, and INT8~\cite{xi2024jetfire}. Frameworks like Pytorch's AMP (Automatic Mixed Precision)~\cite{ansel2024pytorch} and NVIDIA’s Apex ~\cite{nvidia_apex} automated the use of mixed precision training.
Despite these advances, most mixed precision training frameworks use a uniform precision policy across all layers, overlooking potential efficiency gains from layer-wise precision tuning. 
Recent work has begun to incorporate layer-wise information into training precision policies. FGMP~\cite{hooper2025fgmp} leverages Fisher information to guide mixed-precision quantization, ACCORDION~\cite{agarwal2021adaptive} uses Hessian-based analysis to detect critical regimes, and Egeria~\cite{wang2023egeria} introduces a plasticity metric to measure divergence in intermediate activations between a training model and its reference counterpart. To the best of our knowledge, \THISWORK\ is the first to enable sensitivity-aware precision selection for LLM pretraining by jointly considering both loss divergence and weight divergence, and to demonstrate its effectiveness at scale from 1B to 70B models.

{
\noindent\textbf{{Adaptive Mixed Precision for LLM Inference.}
}
Beyond uniform-precision training, several works have explored adaptive or layer-wise mixed precision for LLM inference. Methods like LLM-MQ~\cite{li2023llm}, SliM-LLM~\cite{huang2024slim}, MixLLM~\cite{zheng2024mixllm},  AptQ~\cite{guan2024aptq}, and RaanA~\cite{raana} select precision by minimizing local layer quantization error or consider the impact on loss in the forward pass only, but neglect the cumulative impact of quantization on overall training loss and weight updates. In contrast, \THISWORK\ targets pretraining, where weight dynamics are stronger and higher precision is often required for stability.

{\noindent\textbf{{FP4 LLM Training and Inference.}} 
Building on the success of FP8 pretraining~\cite{liu2024deepseek}, recent studies have begun exploring FP4~\cite{abecassis2025pretraining} as the next frontier for reducing training cost while maintaining stability and accuracy, enabled by NVIDIA Blackwell’s native FP4 support.  Prior studies examine quantization recipes such as block size and rounding~\cite{chmiel2025fp4,tseng2025training,yang2025empirical}, as well as algorithmic enhancements including random Hadamard transforms (RHT)~\cite{tseng2025training} and differentiable gradient estimators (DGE)~\cite{wang2025optimizing} to improve FP4 training accuracy. Despite these advances, a noticeable gap remains between native FP4 training and high-precision baselines (e.g., BF16), suggesting the need for finer-grained mixed-precision approaches. However, FP4 training still lags behind high-precision baselines (e.g., BF16), motivating finer-grained mixed-precision approaches. Complementary work on FP4 inference, such as attention smoothing~\cite{zhang2025sageattention3} and outlier handling~\cite{lee2025amxfp4}, is orthogonal and can be combined with \THISWORK\ for further gains.

\noindent\textbf{Quantization for LLM Inference and Finetuning.}
Quantization has been widely explored for LLM inference to reduce memory and computation costs, leveraging finer quantization granularity, specialized data formats, and outlier suppression techniques. These efforts primarily fall into two categories: post-training quantization (PTQ)~\cite{dettmers2022llm,xiao2023smoothquant, yuan2023rptq, wei2023outlier,lin2024awq, shao2023omniquant,wang2024q,kim2023squeezellm,huang2024billm,ma2024affinequant}, which applies quantization to pretrained models without retraining, and quantization-aware training (QAT)~\cite{liu2023llm, shen2024edgeqat, chen2024efficientqat}, which retrains models to adapt to quantized inference. Additionally, parameter-efficient fine-tuning (PEFT) techniques such as LoRA~\cite{hu2021lora} have been combined with quantization in fine-tuning approaches like QLoRA~\cite{dettmers2024qlora}.
These inference-stage quantization techniques do not directly apply to LLM pretraining because they have substantial overhead and do not consider the overall training quality.
Orthogonal to these, various works propose new quantization formats (e.g., LUQ~\cite{chmiel2023accurate}), whereas \THISWORK\ selects among existing formats (e.g., FP4/FP8) to balance training efficiency and accuracy, and remains compatible with emerging ones.

% These inference-stage quantization techniques do not directly apply to LLM pretraining due to key differences: (1) Training requires higher precision for stable convergence, while inference can use lower precision; (2) Training updates weights every iteration, requiring quantization for every iteration, whereas inference quantizes once and reuses weights; (3) Training involves dynamic activation, gradient, and weight distributions, making a fixed quantization policy ineffective. These challenges highlight the need for quantization strategies for LLM pretraining.

% \input{7.5-future_work}
\section{Conclusion}
We propose \THISWORK, a layerwise mixed-precision framework that introduces a new level of granularity in precision selection per linear layer. By quantifying training quality loss through loss divergence and weight divergence, we formulate per-layer quantization as an Integer Linear Programming (ILP) problem to achieve optimal precision assignments. \THISWORK\ balances training quality and efficiency while seamlessly integrating into LLM training pipelines.
Experiments across 1B, 3B, 7B, and 70B models and different training phases demonstrate that \THISWORK\ outperforms other heuristic-based approaches, achieving up to 80\% FLOPs efficiency savings in FP4 while maintaining accuracy close to the BF16 baseline.

% \begin{acks}
% ...
% \end{acks}

\bibliographystyle{ACM-Reference-Format}
\balance
\bibliography{ref}

\appendix

\newpage 

\section{Proof of the Theoretical Results}

\subsection{Proof of Theorem \ref{thm:perturb}}

Throughout this proof we use ``absolute constant'' to indicate a constant value that does not depend on any variables used in the theorem.

Notice that $\mathbb E \|\delta\|^2 = \epsilon^2$. Let event $\mathcal E = \left\{  C^{-1} \epsilon \leq \|\delta\| \leq C \epsilon \right\}$, where $C > 1$ is an absolute constant. From the known concentration property of Gaussian random vectors, there exists a $C$ such that $\mathbb P(\mathcal E) \geq 0.995$. In the following we condition on all of our statements on $\mathcal E$.

Let $G = \nabla_x g(x) \in \mathbb R^{m \times d}$. Since $g$ is smooth, a small perturbation can be approximated by the first order Taylor expansion. Specifically, we have \begin{align*}
\left\|g(x + \delta) - g(x)\right\|_F & = \left\| G\delta + O\left(S \|\delta\|^2\right)\right\|_F
\\ & \leq  \left\| G\delta \right\|_F + O\left(S \epsilon^2\right).
\end{align*}
Since we assumed $S \ll \epsilon^{-2}$, the second term is negligible, and thus we focus on the first term. 

Let the SVD of $G$ be $G = U\Sigma V$, where $U \in \mathbb R^{m\times m},  V \in  \mathbb R^{d\times d}$ be orthonormal matrices, and $\Sigma \in \mathbb R^{m \times d}$ be a concatenation of a $\min(d,m)$-diagonal matrix and an all-zero matrix. We have \begin{align*}
\|G\delta\|_F = \|U\Sigma V\delta\|_F = \left\|\Sigma \bar \delta\right\|_F,
\end{align*}
where $\bar \delta = V\delta$ which is known to have the same distribution as $\delta$. Let $\sigma = \left\{ \Sigma_{i,i} \right\}_{i=1}^{d}$ be the diagonal entries of $\Sigma$ (if $m < d$, we pad $\sigma$ with $0$ to make it a $d$-dimensional vector). We have \begin{align*}
\left\|\Sigma \bar \delta\right\|_F = \sqrt{ \sum_{i=1}^d \sigma_i^2 \bar \delta_i^2}.
\end{align*}

Since $\bar \delta$ is identically distributed with $\delta$, we have $\mathbb E\bar \delta_i^2 = \frac{\epsilon^2}{d}$. Moreover, since orthonormal transformations does not change the Frobinius norm, we have $\|\Sigma\|_F = \|G\|_F$. Therefore \begin{align*}
\mathbb E \left\|\Sigma \bar \delta\right\|_F^2 = \sum_{i=1}^d \sigma_i^2 \mathbb E\bar \delta_i^2 = \frac{\epsilon^2}{d} \|\Sigma\|_F^2 = \frac{\epsilon^2}{d} \|G\|_F^2.
\end{align*}

From Bernstein's inequality~\cite{vershynin2018high}, we know the value of $\left\|\Sigma \bar \delta\right\|_F^2$ is concentrated around its expectation. Specifically,  there exists an absolute constant $C_2$ such that\begin{align*}
\mathbb P\left\{ \left| \left\|\Sigma\bar \delta\right\|_F^2 - \frac{\epsilon^2}{d}\|G\|_F^2\right| > t \right\} & \leq 2\exp\left( -\frac{C_2 t^2}{ \frac{\epsilon^4}{d^2}\sum_{i=1}^d \sigma_i^4 + t\cdot\frac{\epsilon^2}{d} \max_{i=1}^d \sigma_i^2  } \right)
\\ & \leq 2\exp\left( -\frac{C_2 t^2}{ \frac{\epsilon^4}{d^2}\|G\|_F^4 + t\cdot\frac{\epsilon^2}{d} \|G\|_F^2  } \right).
\end{align*}
It is easy to see that there exists an absolute constant $C_3$, such that with probability at least $0.995$, we have \begin{align}
\left\| G\delta \right\|_F = \left\|\Sigma\bar \delta\right\|_F \leq C_3\cdot \frac{\epsilon}{\sqrt{d}}\|G\|_F.
\end{align}

Since our statement is conditioned on $\mathcal E$, the probability of the overall statement is at least $$1 - \mathbb P\left(\mathcal E^c\right) - \mathbb P\left\{\left\| G\delta \right\|_F  > C_3\cdot \frac{\epsilon}{\sqrt{d}}\|G\|_F\right\} \geq 0.99.$$

\subsection{Proof of Theorem \ref{thm:est-gradient-norm}}

WLOG we only need to prove the theorem for $m = 1$, since in the case of $m > 1$ we only need to apply the 1-d conclusion to each output entry.

Let $Y \sim \mathcal N(0, I_d)$ be a $d$-dimensional standard Gaussian vector. We have $\delta_\epsilon$ is identically distributed with $\epsilon Y$. Using the dominated convergence theorem and the fact that continuous operations commute with limit, we only need to prove the eq.(\ref{eq:est-grad-norm}) with the right-hand side being replaced with \begin{align*}
\mathbb E \left|\lim_{\epsilon \to 0}  \frac{ g(x) - g(x+ \epsilon Y)}{\epsilon}\right|^2.
\end{align*}

Let $G \in \mathbb R^d$ be the gradient of $g$ at point $x$. Notice that $D_Y(x) = \lim_{\epsilon \to 0} \frac{g(x) - g(x+\epsilon Y)}{\epsilon}$ is the directional derivative of $g$ at point $x$ in the direction of $Y$. Since $g$ is smooth, we have $D_Y(x) = \left<G,Y\right>$. Thus we have \begin{align*}
\mathbb E \left|\lim_{\epsilon \to 0}\frac{g(x) - g(x + \epsilon Y)}{\epsilon}\right|^2 & = \mathbb E \left<G, Y\right>^2
\\ & =  \sum_{i=1}^d G_i^2 \mathbb E Y_i^2
\\ & = \sum_{i=1}^d G_i^2 
\\ & = \|G\|^2.
\end{align*}

\end{document}